\documentclass[
tightenlines,
11pt,
pra,
aps,
longbibliography,
onecolumn,
showpacs,
nofootinbib,
superscriptaddress,
notitlepage]{revtex4-2}

\usepackage{microtype}
\usepackage[margin=1in, includefoot, footskip=30pt]{geometry}
\usepackage{amsmath}
\usepackage{amssymb,bm}
\usepackage{amsthm}
\usepackage{braket}
\usepackage{xfrac}
\usepackage[colorlinks, allcolors=black]{hyperref}
\usepackage[capitalize]{cleveref}
\usepackage{subfiles}
\usepackage{graphicx}
\usepackage[font=scriptsize]{caption}
\usepackage[font=scriptsize]{subcaption}
\usepackage{enumitem}
\usepackage{marginnote}
\usepackage{appendix}
\usepackage{booktabs}
\usepackage{algpseudocode}
\usepackage{algorithm}

\theoremstyle{plain}
\newtheorem{thm}{Theorem}[section]
\newtheorem{lem}[thm]{Lemma}
\newtheorem{prop}[thm]{Proposition}
\newtheorem*{cor}{Corollary}

\theoremstyle{definition}

\newtheorem*{ex}{Example}
\newtheorem{condition}{Condition}

\theoremstyle{remark}
\newtheorem*{rem}{Remark}

\usepackage{float}
\usepackage{newfloat}
\DeclareFloatingEnvironment[
    fileext=los,
    listname={List of Algorithms},
    name=Algorithm,
    placement=tbhp
]{alg}

\crefname{thm}{Theorem}{Theorems}
\crefname{lem}{Lemma}{Lemmas}
\crefname{prop}{Proposition}{Propositions}
\crefname{cor}{Corollary}{Corollaries}
\crefname{proof}{Proof}{Proofs}
\crefname{ex}{Example}{Examples}
\crefname{condition}{Condition}{Conditions}
\crefname{rem}{Remak}{Remarks}
\crefname{figure}{Figure}{Figures}
\crefname{alg}{Algorithm}{Algorithms}
\crefname{appsec}{Appendix}{Appendices}

\DeclareMathOperator{\tr}{Tr}
\DeclareMathOperator{\argmin}{argmin}
\DeclareMathOperator{\argmax}{argmin}
\renewcommand{\epsilon}{\varepsilon}
\newcommand{\st}{\gamma}

\newcommand{\temp}{v}

\setlist[enumerate]{wide=\parindent}
\begin{document}

\title{Quantum Algorithms for Structured Prediction}

\author{Behrooz Sepehry}
\affiliation{1QB Information Technologies (1QBit), Vancouver, BC}

\author{Ehsan Iranmanesh}
\affiliation{1QB Information Technologies (1QBit), Vancouver, BC}

\author{Michael P. Friedlander}
\affiliation{1QB Information Technologies (1QBit), Vancouver, BC}
\affiliation{University of British Columbia, Vancouver, BC}

\author{Pooya Ronagh}
\email[Corresponding author:\ ]{pooya.ronagh@1qbit.com}
\affiliation{1QB Information Technologies (1QBit), Vancouver, BC}
\affiliation{Institute for Quantum Computing, Waterloo, ON}
\affiliation{University of Waterloo, Waterloo, ON}

\date{\today}
\begin{abstract}
We introduce two quantum algorithms for solving structured prediction problems.
We first show that a stochastic gradient descent that uses the quantum minimum
finding algorithm and takes its probabilistic failure into account solves the
structured prediction problem with a runtime that scales with the square root of
the size of the label space, and in $\widetilde O\left(1/\epsilon\right)$ with
respect to the precision, $\epsilon$, of the solution. Motivated by robust
inference techniques in machine learning, we then introduce another quantum
algorithm that solves a smooth approximation of the structured prediction
problem with a similar quantum speedup in the size of the label space and a
similar scaling in the precision parameter. In doing so, we analyze a variant of
stochastic gradient descent for convex optimization in the presence of an
additive error in the calculation of the gradients, and show that its
convergence rate does not deteriorate if the additive errors are of the order
$O(\sqrt\epsilon)$. This algorithm uses quantum Gibbs sampling at temperature
$\Omega (\epsilon)$ as a subroutine. Based on these theoretical observations, we
propose a method for using quantum Gibbs samplers to combine feedforward neural
networks with probabilistic graphical models for quantum machine learning. Our
numerical results using Monte Carlo simulations on an image tagging task
demonstrate the benefit of the approach.
\end{abstract}

\maketitle

\section{Introduction}
\label{sec:intro}

Structured prediction is an area of machine learning, where the aim is to learn
an association of the input data to a structured output
\cite{nowozin2014advanced}. Structured prediction tasks arise naturally in many
real-world applications. For example, predicting temporal structures (e.g., the
parse tree of a sentence, or the coreferences between nouns and pronouns in
texts) are important in natural language processing applications
\cite{daume2006practical}. Spatial structures (e.g., the segmentation of an
image into meaningful components, the geometry of molecules) are of
interest in several areas of application, such as image processing, computer
vision, and computational biology \cite{jiang2020spatial}.

The structues of interest in structured prediction are often discrete and
combinatorial in nature. As a result, while a description of an admissible
structure can be represented efficiently, the number of possible valid
structures is exponentially larger than these representations. As such, viewing
structured prediction tasks as classification problems results in exponentially
large numbers of labels. This is why classical machine learning has to employ
techniques such as generalization of support vector machines (SVM) to structured
SVMs (SSVM) \cite{yu2011improved} and generative models \cite{sohn2015learning}
to solve structured prediction problems.

The inherently combinatorial nature of structured prediction tasks gives rise to
piecewise smooth models, as in the case of SVMs. However, predicting a
structured output involves the prediction of vectors, rather than simply scalar
value assignments. Therefore, the combinatorial aspect of the task is especially
egregious in structured prediction. For example, in structured SVMs (SSVM), the
number of pieces in the piecewise smooth model is often exponentially large in
terms of the dimension of prediction vectors.

The underlying optimization problem in structured prediction tasks is of the
form
\begin{equation}\label{eq:objective}
  \min_w\ \frac{1}{n}\sum_{i=1}^{n} g_i(w),
  \quad\mbox{where}\quad
  g_i(w) = \max_{y\in \mathcal Y} f_i(y, w)\,,
\end{equation}
where $f_i$ are strongly convex with Lipschitz continuous gradients,
and $\mathcal Y$ is a finite set. This can be easily extended to the case
in which each function $f_i$ is defined on a distinct domain $\mathcal Y_i$.
The size of $\mathcal Y$ can cause the evaluation of the $\max$ operator
to be computationally intractable. In SVMs and SSVMs, $w$ represents the
trainable weights of the model, the functions $f_i$ represent the margins of
the points in a dataset from classifying hyperplanes with additional
regularizer terms, and $\mathcal Y$ is the exponentially large label set
discussed above. Problems in the form of \eqref{eq:objective} arise
in other machine learning applications as well.

In this paper, we present two methods for solving the above optimization
problem following the approaches taken in classical convex optimization for
nonsmooth optimization. The first approach relies on \emph{subgradient methods},
which are among the common techniques for optimization of nonsmooth models.
We introduce a quantum algorithm based on a subgradient method
and the quantum minimum finding algorithm \cite{durr1996quantum}.
Another common technique for dealing with nonsmooth models is \emph{smooth
approximation}, for example, by using softmax operators. Although
these techniques are effective at hiding the nonsmooth aspects of the problem by
replacing a piecewise smooth problem with a single smooth approximation,
computing that approximation can be intractable when the
number of pieces is large. In our second quantum algorithm, we consider a smoothing
that combines softmax approximation and quantum Gibbs sampling.
The quantum Gibbs sampler is used to estimate expected values of certain
observables of the Boltzmann distribution of a many-body system, which in turn
provide estimations of gradients of the smooth approximation.

The performance advantages of the quantum algorithms introduced in this paper
are as follows.
\begin{enumerate}
\item[(a)]
Both of our quantum algorithms achieve quadratic quantum speedups (up to
polylogarithmic factors) in terms of the size of the label space in structured
prediction tasks. This is an important speedup for machine learning applications
since the techniques for classification with a small number of labels do not
translate into performant methods in tasks with large numbers of labels
\cite{bi2013efficient}.

\item[(b)]
From the machine learning point of view, optimizing the smooth approximation of
the objective function in \eqref{eq:objective} itself is of natural interest. The
softmax approximation allows for the optimizer to fit a reasonable model to the
structured prediction problem while avoiding settling into erroneous minima that
are artifacts of limited training data. This may result in better generalization
and more-robust learning \cite{lee2001ssvm}.

\item[(c)]
Quantum algorithms that achieve quadratic speedups
for classical optimization problems often do so at the expense of much worse
scaling in terms of the solution precision they achieve. For example, quantum
algorithms for linear programming (LP) and semidefinite programming (SDP) incur
higher-order polynomial complexities with respect to solution precision (see
\cref{sec:related-lit}). The higher-order polynomial
scalings in the precision of the solution are obstacles to the practical
applicability of these algorithms. In contrast, our quantum algorithms possess
linear scalings (up to polylogarithmic factors) in the precision of the
solutions they return. This scaling agrees with the classical optimal bounds for
first-order methods in the convex optimization of nonsmooth functions
\cite{shamir2013stochastic, nesterov2005smooth}.
\end{enumerate}

\subsection{Related literature}
\label{sec:related-lit}

Stochastic gradient descent (SGD) is a simple and efficient algorithm that has
become the core algorithm used in classical large-scale convex optimization and
its applications in machine learning. SGD relies on classical queries to an
unbiased estimator of the gradients of the objective function. SGD generalizes
to not-differentiable convex functions, in which case it suffices to have access
to unbiased estimators of the subgadients (i.e., the expected queried vector has
to be an element of the subgradient set). SGD can only achieve a
$\widetilde O(1/\epsilon^2)$ convergence rate for nonsmooth functions
\cite{shamir2013stochastic}, making it provably suboptimal for nonsmooth
optimization \cite{harvey2018tight}. However, variants of it, such as SGD with
\emph{suffix averaging} \cite{rakhlin2012making} or SGD with
\emph{polynomial-decay averaging} \cite{shamir2013stochastic,lacoste2012simpler}
(SGDP), achieve the optimal convergence rate of $\widetilde O(1/\epsilon)$.

In this paper, we use stochastic (sub-)gradient descent with polynomial-decay
averaging (SGDP) to solve the structured prediction problem
\eqref{eq:objective}. We use the
quantum minimum finding algorithm (QMF) of D\"urr and
H{\o}yer \cite{durr1996quantum} as a subroutine to solve the inner discrete
optimization problem in \eqref{eq:objective} over the label space $\mathcal Y$.
QMF is based on Grover's search algorithm \cite{grover1996fast}. Since QMF has a
randomized nature, it may fail to return a correct unbiased estimator of the
subgradient in \eqref{eq:objective}. Our analysis shows that the failure of
QMF can be overcome if its failure rate is kept at $O(\epsilon)$.

A second approach to solving problems in the form of \eqref{eq:objective} is to
approximate the piecewise smooth problem with a fully smooth one and use
variants of gradient descent design for smooth problems, such as SAGA
\cite{defazio2014saga}. Each
function $g_i$ is replaced with an approximation that is strongly convex with
a Lipschitz continuous gradient. However, these smooth approximations typically
rely on replacing the $\max$ operator with the differentiable softmax operator
\cite{gao2017properties,beck2012smoothing}, that is, each function $g_i$ is
replaced by the smooth approximation
\begin{align}
g_i^\beta(w) := \frac{1}\beta\log\sum_{y\in\mathcal Y}e^{\beta f_i(y,w)}\,,
\end{align}
which is at least as computationally difficult as evaluating the original max
operator. This approximation can be interpreted from a
thermodynamic perspective, wherein each $g_i^\beta$ represents the free energy of a
system with an energy spectrum described by $f_i$. This motivates the use of
quantum Gibbs samplers for computing such smooth approximations.

It has been speculated for the past 20 years that quantum computers can be used to
generate samples from Gibbs states \cite{terhal2000problem}. Since then,
many algorithms for Gibbs sampling based on a quantum-circuit model have been
introduced
\cite{poulin2009sampling,temme2011quantum,kastoryano2016quantum,chowdhury2016quantum,van2017quantum}.
The Gibbs sampler of
van Apeldoorn et al. \cite{van2017quantum}, has a logarithmic dependence on the
error of the simulated distribution. The sampler of Chowdhury and Somma
\cite{chowdhury2016quantum} similarly has a logarithmic error dependence, but
must assume a query access to the entries of the square root of the problem
Hamiltonian. These quantum-circuit algorithms use phase estimation and
amplitude amplification techniques to create a quadratic quantum speedup in
Gibbs sampling.

Moreover, numerical and experimental heuristics may provide even better
practical performance for Gibbs sampling. The Gibbs sampler of
Temme~et~al.~\cite{temme2011quantum} has an
unknown runtime, but has the potential to provide efficient heuristics since it
relies on a quantum Metropolis algorithm.
Experimentally, quantum and semi-classical evolutions can be used as
physical realizations of improved Gibbs samplers. For example, contemporary
investigation in quantum adiabatic theory focuses on adiabaticity in open
quantum systems
\cite{sarandy2005adiabatic,avron2012adiabatic,albash2012quantum,bachmann2016adiabatic,venuti2016adiabaticity}.
These authors prove adiabatic theorems to various degrees of generality and
assumptions. These adiabatic theorems suggest the possibility of using
controlled adiabatic evolutions of quantum many-body systems as samplers of the
instantaneous steady states of quantum systems.
Takeda~et~al.~\cite{takeda2017boltzmann} show that a network of non-degenerate optic
parametric pulses can produce good estimations of Boltzmann distributions.
Another possible approach to improved Gibbs samplers is to design customized
Gibbs sampling algorithms that rely on Monte Carlo and quantum Monte Carlo
methods implemented on digital high-performance computing hardware
\cite{matsubara2017ising,okuyama2017ising}.

Boltzmann distributions arise naturally in machine learning computations due to
the principle of maximum entropy. Sampling from Boltzmann distributions,
although computationally challenging, is unavoidable in Markovian models of
reasoning. For instance, we refer the reader to
Hammersley--Clifford theorem \cite{wainwright2008graphical,koller2009probabilistic}
in the context of training undirected probabilistic graphical models (UGM).
With the success of deep neural networks in many practical applications,
improving their performance by combining them with UGMs
has become an active area of research. For example, a combination of
convolutional neural networks (CNN) and fully connected
conditional random fields (a type of UGM), has achieved state of the art
performance in image segmentation---a critical task in computer vision.
In image tagging, another computer vision task, combining CNNs
with Ising models (as a UGM with binary random variables and pairwise interactions),
has been found to provide improved accuracy \cite{chen2015learning}.
\cite{kim2011higher}
formulates image segmentation as a correlation clustering problem and solve
it using UGMs with higher-order interactions.
Similarly \cite{yu2009learning} solves the natural language processing
problem of noun phrase coreference by formulating it as a correlation clustering
problem.

The idea of using Gibbs sampling as a subroutine in quantum machine learning has
already been considered. Wiebe et al.~\cite{wiebe2014quantum} use Gibbs state
preparation to propose an improved framework for quantum deep learning. Crawford
et al.~\cite{crawford2016reinforcement} and Levit et al.~\cite{levit2017free}
introduce a framework for reinforcement learning that uses Gibbs states as
function approximators in \emph{Q}-learning. Quantum Gibbs sampling has recently
been shown to provide a quadratic speedup in solving linear programs (LP) and
semi-definite programs (SDP) \cite{brandao2017quantum,brandao2017quatumsdp,van2017quantum}.
The speedup in these quantum algorithms with respect to the
problem size often comes at the expense of much worse scaling in terms of
solution precision. For example, van Apeldoorn~et~al.~\cite{van2017quantum}
propose a quantum algorithm for LP that requires $\widetilde O(\epsilon^{-5})$
queries to the input of the LP, and an algorithm for SDPs that requires $\widetilde
O(\epsilon^{-8})$ queries to the input matrices of the SDP, where $\epsilon$ is
an additive error on the accuracy of the final solution.
Van~Apeldoorn and Gily{\'e}n \cite{van2018improvements} later improved the
scaling of their result by further analysis and reduced the dependence on
precision parameters to $\widetilde O(\epsilon^{-4})$ and more recently to
$\widetilde O(\epsilon^{-3.5})$ \cite{van2019quantum}. Several lower bounds proven
in \cite{van2017quantum,van2018improvements} suggest that these results cannot
be improved significantly further. In particular, the polynomial dependence on
precision parameters is necessary. In contrast, the quantum algorithms proposed
in this paper provide (optimal) $\tilde O(\epsilon^{-1})$ scaling in precision
of the solutions they return.

\subsection{Summary of results}
\label{sec:summary-results}

The two classical optimization approaches discussed above for (a)
solving the original min-max problem directly with subgradient method, and (b)
solving a smooth approximation of it using SAGA \cite{defazio2014saga}, inspired
the design and analysis of the two quantum algorithms we present in this paper.
Our contributions are summarized as follows. The proofs of all propositions can
be found in the appendix (\cref{sec:appendix}).

{\noindent\vskip1mm\hskip-5.4mm \bf \cref{thm:SGDP}\,.}
We show that stochastic subgradient descent with polynomial-decay averaging
(SGDP), under conditions that take the probabilistic errors of quantum minimum
finding into account, can solve the optimization problem \eqref{eq:objective}
with the same optimal convergence rate as SGDP under its original conditions.

{\noindent\vskip1mm\hskip-5.4mm \bf \cref{thm:qSGDP}\,.}
We then derive the query complexity of Q-SGDP (SGDP using quantum minimum
finding) and observe a quadratic speedup in terms of the size of the discrete
set $\mathcal{Y}$. The caveat, however, is that the query complexity reported in
this theorem has a factor of $\log \frac{1}G$ where $G$ is the minimum gap
attained by the functions $f_i(y,w)$ as functions of $y$ as $w$ ranges over its
various observed values throughout the algorithm.

{\noindent\vskip1mm\hskip-5.4mm \bf \cref{thm:asaga_complexity_1}\,.}
We show that SAGA, under conditions that take the probabilistic errors of
quantum Gibbs sampling into account, minimizes a smooth approximation of the
objective function in \eqref{eq:objective} with the same optimal convergence
rate as SAGA under its original conditions.

{\noindent\vskip1mm\hskip-5.4mm \bf \cref{thm:qsaga-original}\,.}
We then derive the query complexity of Q-SAGA (SAGA using quantum Gibbs
sampling) for solving the original nonsmooth optimization problem
\eqref{eq:objective}. We conclude that Q-SAGA also achieves a quadratic speedup
in terms of $|\mathcal{Y}|$ without a depedence on the minimum gap,
at the expense of a slightly worse scaling in terms of $\epsilon$.

{\noindent\vskip1mm\hskip-5.4mm \bf Experiments of \cref{sec:image-tagging}\,.}
To show the real-world applicability of our approach, we formulate image tagging
as a structured prediction task. We then train a neural network with leading
deep layers and a trailing probabilistic graphical model for this task. Our
numerical results show that this hybrid architecture trained with a structured
prediction objective function in the form of smooth approximation of the
objective in \eqref{eq:objective}, can outperform a purely deep model with the
same number of parameters.

\section{Background}
\label{sec:background}

We first present a brief account of SVMs and SSVMs.
We refer the reader to \cite{ng2010support} for the
basics of SVMs and to \cite{yu2011improved} for SSVMs. We then introduce
the more general framework of structured prediction tasks in machine learning.
These models are of particular interest in scenarios where the numbers of
labels is very large, for example, when a label can be any of the exponentially
many binary vectors of a given dimension.

\subsection{SVMs and SSVMs}
\label{subsec:ssvm}

Let $\mathcal{X}$ be a feature set and $\mathcal{Y} = \{-1, 1\}$ be the label
set. We are also given a training dataset
$\mathcal{S} \subseteq \mathcal{X} \times \mathcal{Y}.$
A linear classifier is then given by two (tunable) parameters $w$ and $b$
defining a separating hyperplane $w^T x + b$. For a point $(x, y) \in \mathcal S$,
the positivity of $y(w^T x + b)$ indicates the correct classification of $x$.
The SVM optimization problem can be expressed as
\begin{align}
\begin{split}
\min_{w, b} \quad& \frac{1}2\|w\|^2  \\
\text{s.t.} \quad& y \left(w^T x + b\right) \geq 1
\quad \forall (x, y) \in \mathcal{S}.
\end{split}
\end{align}
The constraints ensure not only that every $(x, y) \in \mathcal S$ is classified
correctly, but done so with a confidence margin. If $y (w^T x + b)$ is positive,
one could superficially satisfy $y \left(w^T x + b\right) \geq 1$ by scaling
up $w$ and $b$. To avoid this we minimize the objective $\frac{1}2 \|w\|^2$. In
other words, the constraints ensure that the distance of $\mathcal S$ to the
classifying hyperplane is at least $1/\|w\|$, and the objective function asks
for this margin to be maximized.

Often, the above optimization problem is infeasible, so we would rather solve a
relaxation of it by introducing slack variables for every data point in
$\mathcal{S}$:
\begin{align}
    \begin{split}
        \min_{w, b, \xi} \quad& \frac{\lambda}2  \|w\|^2
        + \sum_{(x, y) \in \mathcal{S}} \xi_{(x, y)} \\
        \text{s.t.} \quad& y \left(w^T x + b\right) \geq 1 - \xi_{(x, y)}
        \quad \forall (x, y) \in \mathcal{S}\\
        & \xi_{(x, y)} \geq 0 \quad \forall (x, y) \in \mathcal{S}.
    \end{split}
\end{align}
Here $\lambda$ is a hyperparameter adjusting the relative importance of
the two terms in the objective.

For simplicity, we will remove the bias $b$ from the rest of the analysis and
consider it a trainable feature of $x$.
Let $\mathcal{Y}$ now contain more than just two
labels. The score of class $y$ is then represented by the dot product
$w_y^Tx$. The Crammer--Singer formulation of the multi-label SVM problem is the
following:
\begin{align}
\begin{split}
\min_{w, \xi} \quad& \frac{\lambda}2
\sum_{y \in \mathcal{Y}} \|w_y\|^2+ \sum \xi_{(x, y)} \\
\text{s.t.} \quad& w_{y}^T x - w_{y'}^T x \geq 1 - \xi_{(x, y)} \quad
\forall (x, y) \in \mathcal{S}, \forall y' \in \mathcal{Y} \setminus \{y\}\\
& \xi_{(x, y)} \geq 0 \quad \forall (x, y) \in \mathcal{S}.
\end{split}
\end{align}
We can rewrite this in a notation more suitable for introducing SSVMs as a
generalization of SVMs. We first concatenate the weight vectors $w_y$ into a
single vector $w^T = (w_1^T, \ldots, w_k^T)$. We then introduce the
\emph{joint feature map} $\Phi(x, y)= (0, \ldots, x, \ldots, 0)$,
with $x$ being in the $y$-th position and all other elements 0. Lastly,
we introduce a notion of \emph{distance} or \emph{loss function} on $\mathcal{Y}$:
\begin{align}
\Delta(y', y) = \begin{cases} 1 & y = y'\,,\\ 0 & \text{otherwise}. \end{cases}
\end{align}
Then, the model can be rewritten as
\begin{align}
\begin{split}
\min_{w, \xi} \quad& \frac{\lambda}2  \|w\|^2+ \sum \xi_{(x, y)}  \\
\text{s.t.} \quad& \xi_{(x, y)}
\geq \Delta(y', y) - w^T \Phi(x, y) + w^T \Phi(x, y')
\quad \forall (x, y) \in \mathcal{S}, \forall y' \in \mathcal{Y}\\
& \xi_{(x, y)} \geq 0 \quad \forall (x, y) \in \mathcal{S}.
\end{split}
\label{eq:ssvm_constrained}
\end{align}

The above model is that of an SSVM in general, with possibly more-complicated
joint feature maps $\Phi$ and loss functions $\Delta$. This optimization
problem can be written in the unconstrained form of $\min_w f_{\text{SSVM}}(w)$
for
\begin{align}
\label{eq:ssvm_unconstrained}
f_{\text{SSVM}}(w)= \frac{\lambda}{2}  \| w\|^2
+ \sum_{(x,{y})\in \mathcal{S}} \max_{y'}& \left\{ \Delta(y', {y})
+ w^T \left[\Phi(x,y') - \Phi(x,{y})\right] \right\},
\end{align}
which is a min-max optimization problem of the form
\begin{align}
\label{eq:ssvm_dual}
\min_{w} \left( f(w)=
\left\{ \sum_{(x,{y})\in \mathcal{S}} \max_{y'} f_{(x, y)} (y'; w)\right\}\right)\,,
\end{align}
where the summands $f_{(x, y)}(y'; w)$ are of the form
\begin{align}
\label{eq:ssvm_obj}
f_{(x, y)}(y'; w)= \frac{\lambda}{2|\mathcal S|} \|w\|^2
+ \Delta(y', y) + w^T \left[\Phi(x,y') - \Phi(x,{y})\right].
\end{align}

Without the regularizer term,
problem \eqref{eq:ssvm_constrained} is therefore readily of the mathematical
form of the Lagrangian dual problems studied in
\cite{ronagh2016solving,karimi2017subgradient},
and cutting plane or subgradient methods could
be used to solve them efficiently under the assumption of the existence of \emph
{noise-free} discrete optimization oracles.
It is also a linear problem, and the quantum linear programming
technique of \cite{van2017quantum} could be used to provide quadratic speedup
in the number of constraints and variables of the problem. In most practical
cases, however (see below), the instances are very large, and it would not be
realistic to assume the entire problem is available via an efficient circuit
for oracle construction. Stochastic gradient descent methods overcome this
difficulty (for classical training data) by randomly choosing training samples
or mini-batches. This is also our approach in what follows.

\subsection{Structured Prediction}
\label{subsec:structured-prediction}

We now introduce the general framework of \emph{structured prediction} as a
supervised learning task in machine learning. SSVMs are only one of the
mathematical models used to solve structured prediction problems. As we will see,
the distinguishing factor between techniques for solving structured
prediction problems is the choice of an objective function similar to
\eqref{eq:ssvm_obj}. We will assume that
structured prediction problems are equipped with the following.

\begin{enumerate}
\item[(a)] A training dataset
$\mathcal{S}\subseteq \mathcal{X} \times \mathcal{Y}$.

$\mathcal{X}$ and $\mathcal{Y}$ are, respectively, the set of
all possible inputs and outputs.
The elements of $\mathcal Y$ encode a certain \emph{structure} (e.g., the
syntactic representation of an English sentence). In structured prediction, the
outputs are therefore vectors instead of scalar discrete or real values.
In particular, the set $\mathcal{Y}$ may be exponentially large in the size of
the input. This distinguishes structured prediction from multi-label
classification.

\item[(b)] A scoring function
$s_w :\mathcal{X} \times \mathcal{Y} \to \mathbb{R}$.

The scoring function $s_w (x, y)= s(x, y; w)$
is indicative of suitability of a label $y$ for a
given input $x$, where $w$ is a vector of tunable parameters.
The \emph{predictor} for the model which is a function mapping the
input $x$ and parameter $w$ to an output prediction $y$ is therefore
\begin{align}
    \label{eq:predictor}
    h_w(x) = \argmax_{y'} s(x, y'; w).
\end{align}

\item[(c)] A real-valued loss function
$\Delta:\mathcal{Y}\times \mathcal{Y}\rightarrow \mathbb{R}$.

The goal is to find a predictor $h_w$ like \eqref{eq:predictor} that minimizes
the \emph{empirical risk}
\begin{align}
    \label{eq:empirical_risk}
    R(h_w) = \frac{1}{|S|} \sum_{(x,y)\in \mathcal{S}} \Delta(h_w(x), y)\,.
\end{align}

We assume that the minimum of $\Delta$ over its first component is uniquely
attained along its diagonal, that is,
\begin{align}
    \label{eq:Delta_argmin}
    y=\argmin_{y'}\Delta(y',y)\,,
\end{align}
and without loss of generality, we may assume that $\Delta$ vanishes on its diagonal
\begin{align}
    \label{eq:Delta_cond_1}
    \Delta(y,y) = 0 \quad \forall y\in \mathcal{Y}\,,
\end{align}
since we can always shift it to $\Delta'(y',y) = \Delta(y',y) - \Delta(y,y)$.
This decreases the empirical risk by the constant
\begin{align}
\frac{1}{|S|}\sum_{(x,y)\in \mathcal{S}}\Delta(y,y),
\end{align}
which is an invariant of $\mathcal{S}$.

\end{enumerate}

\begin{ex}
\label{ex:Ising_model}
In the SSVM framework of \cref{subsec:ssvm}, the loss
function $\Delta$ is simply the Kronecker delta function, in other words
$\Delta(y, y') = \delta_{y, y'}$.
and the scoring function is linear in the
training parameters
$s(x, y; w)= w^T \Phi(x, y)\,.$ Furthermore, if $\Phi(x, y)$ is quadratic in
$y$ and $\mathcal{Y}$ is the set of binary vectors with a fixed length, then
$s(x, y; w)$ corresponds to the energy of an Ising model, which
appears in many structured prediction tasks as discussed in
\cref{sec:intro}. With this scoring function $s$, the task of prediction
using the predictor $h$ corresponds to finding the ground state of the Ising
model.
\end{ex}

\subsection{A Min-Max Optimization Problem}
\label{sec:min_max}

We now present a general mathematical programming model motivated by our machine
learning discussion above and in \cref{sec:nonsmooth_prelude,sec:smooth_prelude}
consider quantum algorithms for solving them.

We define the objective function
\begin{equation}
\label{eq:min_max_func}
  f(w)= \frac{1}{n}\sum_{i=1}^{n} g_i(w),
  \quad\mbox{where}\quad
  g_i(w) = \max_{y\in \mathcal Y} f_i(y, w)\,.
\end{equation}
Here $w$ is a vector of tunable real-valued parameters, $n$ is a positive
integer, and all $f_i$ are strongly convex real-valued functions of
$w$ with Lipschitz continuous gradients.
Furthermore each $f_i$ is defined in its first
argument $y$ over a finite set $\mathcal Y$.
In practical machine learning examples, $f_i$ could
be strongly convex because of addition of a strongly convex regularizer
(e.g., $L^2$ regularizer) to an already convex loss function.
We are interested in solving the optimization problem
\begin{align}
\label{eq:min_max_opt}
w_* = \argmin_w f(w)\,.
\end{align}
Although the functions $f_i$ are differentiable, $f$ is not generally
differentiable because of the $\max$ operator involved. However, since the
$\max$ operator preserves convexity, $f$ is a convex function.

\section{Nonsmooth Optimization}
\label{sec:nonsmooth_prelude}

Here we provide a time complexity analysis on the optimization of problem
\eqref{eq:min_max_func} using a stochastic variant of the subgradient method
\cite{harvey2018tight,shamir2013stochastic} that incorporates a
gradient-averaging scheme known \emph{polynomial-decay averaging}
\cite{shamir2013stochastic,lacoste2012simpler}.

Our approach is to use stochastic (sub-)gradient descent with polynomial-decay
averaging (SGDP) to optimize the nonsmooth objective function $f$
with quantum minimum finding providing the approximations of subgradients for
SGDP. Because the estimates of the subgradients are not exact, we need to
revisit the convergence of SGDP in the presence of errors in calculating the
subgradients and do so in the following sections.

\subsection{A-SGDP: Approximate SGDP}
\label{sec:a-SGDP}

We assume the following condition about the function $f$.

\begin{condition}
\label{cond:rsg_1}
Each function $f_i$ is $\mu$-strongly convex, and as a result
each $g_i$ is also $\mu$-strongly convex. The vector $w$ is restricted to a
convex set $\mathcal{W}$. Furthermore, the subgradients of $f_i(y,w)$ exist
and have
the bounded norms
\begin{align}
\label{eq:rsg_grad_bound}
\sup_{w,i,y} \left\{\|v\|^2: v \in \partial f_i(y,w)\right\} \leq M\,,
\end{align}
where the supremum ranges over every index $i$, every $y \in \mathcal Y$,
and every $w \in \mathcal W$. Finally, each function
$f_i$ has an efficient quantum oracle, that is, one that acts on
$O(\mathrm{polylog} (\frac{1}\delta, |\mathcal Y|))$
qubits to compute $f$ with an additive error of $\delta$.
\end{condition}

The algorithm is as follows. We use the SGDP algorithm of
\cite{shamir2013stochastic}, where, at each iteration, we compute a maximizer $y$
for a function $f_i$ using the quantum minimum finding algorithm
\cite{durr1996quantum}. In the end, we return the weighted
average of $w$ at each iteration according to the polynomial-decay
averaging scheme. SGDP, combined with the quantum minimum finding
algorithm, yields what we refer to as the \emph{quantum} SGDP (Q-SGDP)
algorithm.

Before analyzing Q-SGDP, we introduce and analyze the approximate variant of
SGDP, called \emph{approximate} SGDP (A-SGDP) \cref{alg:A-SGDP}, in which we
account for a probabilistic rate of failure in finding a maximizer for the
discrete optimization of $f_i$ over $\mathcal Y$.

\begin{alg*}
\noindent\framebox{\begin{minipage}[t]{0.986\textwidth}
\label{alg:rsg}\small

\begin{algorithmic}[1]
    \Procedure{A-SGDP}{$w^0\in \mathcal{W}$: initial point,
                       $T \in \mathbb{N}$: number of iterations,
                       $\eta\in \mathbb{R}^+$: SGDP parameter}

    \State $\bar{w}^{0}_{\eta}:= w^0$
    \Comment{Initialize the polynomial-decay averaging value.}
    \For{$t\in \{0, \dots, T-1\}$}
    \State $i \sim \text{Unif}\{1, \dots, n\}$
    \Comment{Draw the index $i$ uniformly at random.}
    \State $\hat{y}_i^t \sim \widehat{\argmax}_{y\in \mathcal{Y}} f_i(y,w^t)$
    \Comment{Estimate ${\argmax}_{y\in \mathcal{Y}} f_i(y,w^t)$
        with at least $1-\zeta$ success probability.}
    \State $\widehat{\partial g_i(w^t)}:= \partial f_i(\hat{y}_i^t, w^t)$
    \Comment{Calculate the gradient.}
    \State $v^{t+1}:= w^t - \gamma_t \widehat{\partial g_i(w^t)}$
    \Comment{Perform a gradient descent step with step size $\gamma_t$.}
    \State $w^{t+1}:= \Pi_{\mathcal{W}}(v^{t+1})$
    \Comment{Project onto the convex set $\mathcal{W}$.}
    \State $\bar{w}^{t+1}_{\eta}
        := \frac{t + 1}{t + \eta + 2} \bar{w}^{t}_{\eta}
         + \frac{\eta + 1}{t + \eta + 2} {w}^{t+1}$
    \Comment{Update the polynomial-decay averaging value.}

    \EndFor
    \State \Return $\bar{w}^T_\eta$
    \EndProcedure
\end{algorithmic}
\end{minipage}}
\caption{Pseudocode for the A-SGDP algorithm.}
\label{alg:A-SGDP}
\end{alg*}

We now analyze the complexity of A-SGDP. Recall that a minimizer of $f$ is
denoted by $w_*$ in \eqref{eq:min_max_opt}.

\begin{thm}
\label{thm:SGDP}
Under \cref{cond:rsg_1}, given a target accuracy $\epsilon > 0$, A-SGDP
finds a point $w\in \mathcal{W}$ satisfying
$|f(w)-f(w_*)| \leq \epsilon$ with a probability of at least $1/2$ in
$T = {O}\left(\frac{M}{\mu \epsilon}\right)$
descent iterations, if the stepsize at iteration $t$ is chosen as
$\gamma^t = \frac{1}{\mu t}$,
and the failure probability in solving $\argmax_y f_i(y,w^t)$
at iteration $t$ is at most $\zeta = \frac{1}{4T}$.
\end{thm}

\subsection{Quantum minimum finding}

The quantum minimum finding algorithm (QMF) \cite{durr1996quantum} relies
on coherent queries to an oracle implementing a real-valued function
$f: \mathcal Y \to \mathbb R$ on a discrete domain $\mathcal Y$ of size $N$. QMF
performs several iterations of Grover's search on a comparative oracle between
the values of $f$ and a classically chosen threshold value to determine a next
point $y \in \mathcal Y$ on which $f$ attains a smaller value than the
threshold. After these iterations, QMF returns an optimizer of $f$ with high
probability using $\widetilde O(\sqrt N)$ queries. The success probability of
QMF can be trivially boosted to any target $1 - \zeta$ by $O(\log(1/\zeta))$
repetitions of the algorithm.

We now provide a statement of the query complexity of QMF following
\cite[Theorem~1]{durr1996quantum}. As we are interested in the complexity of
queries made to the oracles of the functions $f_i$, we take into account
the number of additional elementary gates required to implement the mentioned
comparative oracle. This is also done in \cite[Theorem 49]{van2017quantum}.
\cref{lem:grover_max} is a statement of QMF query complexity using the notation
developed in this paper.

\begin{lem}
\label{lem:grover_max}
Let $i$ be a fixed index and $f_i$ be the corresponding function as defined
in \eqref{eq:objective}. Let $F$ be a bound on the absolute values of $f_i$, that is,
$F = \max_{y\in \mathcal{Y}} |f_i(y,w)|$, and $G$ be the difference between
the maximum value of the function and the second-largest value of $f_i$.
Let $U$ be a unitary that implements $f_i$ and acts on $q$ qubits in order to do so.
There exists a quantum algorithm that returns a (not necessarily unique)
point $\hat{y}_i \in \argmax_{y\in \mathcal{Y}} f_i(y,w)$, with a probability
of at least $1 - \zeta$, in
$O\left(\sqrt{|\mathcal{Y}|} \log(1/\zeta) \right)$
queries to the oracle of $f_i$, and using
$O\left(\sqrt{|\mathcal{Y}|} \log(F/G) \log(1/\zeta) \right)$
additional other quantum gates.
\end{lem}

\subsection{Q-SGDP: Quantum SGDP}

We may now analyze the complexity of Q-SGDP, which is the result of combining
A-SGDP and the quantum minimum finding algorithm. The query complexity of Q-SGDP
can be found using the asymptotic number of descent steps reported in
\cref{thm:SGDP} as follows.

\begin{thm}
\label{thm:qSGDP}
Under \cref{cond:rsg_1}, given a target accuracy $\epsilon > 0$, Q-SGDP
finds a point $w\in \mathcal{W}$ satisfying
$|f(w)-f(w_*) | \leq \epsilon$
with a probability of at least $1/2$ in
$T = \widetilde{O}\left(\frac{M \sqrt{|\mathcal{Y}|}}{\mu \epsilon}
\right)$
queries to the quantum oracles of $f_i$ and using
$\widetilde{O}\left(
\frac{M \sqrt{|\mathcal{Y}|}}{\mu \epsilon} \log \frac{F}{G}\right)$
additional quantum gates. Here $F$ is a bound on the absolute values of $f_i$ for
all $i, y$, and $w$ and $G$ is the minimum gap attained by the functions $f_i$
for different values of $y\in \mathcal{Y}$ throughout the runtime.
\end{thm}

\section{Smooth Optimization}
\label{sec:smooth_prelude}

In \cref{sec:nonsmooth_prelude}, we considered subgradient method for minimizing
the nonsmooth objective function \eqref{eq:min_max_func}. In convex
optimization, smooth approximation of nonsmooth objective functions is a common
method for designing improved gradient-based solvers \cite{beck2012smoothing}.
We construct such a smooth approximation of the function $f$, and find the
minimum of the approximation.

One approach to smoothing the $\max$ of a set of functions is \emph{softmax
smoothing} \cite{beck2012smoothing}. For a finite set $\mathcal Y$ and $\beta > 0$, the
softmax approximation of the $\max$ operator over
a set of values $\mathcal Y$ is defined as
\begin{align}
\label{eq:softmax}
{\max_{y \in \mathcal Y}}^\beta y =
\frac{1}{\beta}\log \sum_{y \in \mathcal Y} \exp(\beta y)\,.
\end{align}
This is the negative free energy of a physical system with an energy spectrum
$\{-y: y \in Y\}$.
We now apply smoothing to the range of every summand $f_i$ in \eqref{eq:min_max_func}
and the resultant summation is called the smooth approximation of $f$ at
inverse temperature $\beta$, denoted by $f^\beta(w)$:
\begin{align}
\label{eq:smooth_min_max_func}
f^\beta(w) =  \frac{1}{n} \sum_i  {\max_{y\in \mathcal{Y}}}^\beta f_i(y, w)\,.
\end{align}
We note that $f^\beta (w)$ converges uniformly to $f(w)$ in
the limit of $\beta \to \infty$
(refer to \eqref{eq:smooth_ineq} below).
So, on one hand, $\beta$ can be interpreted as
the thermodynamic inverse temperature at equilibrium for each energy
function $-f_i$ and, on the other hand, as a parameter controlling the amount of
smoothing imposed on $f$. That is, when $\beta$ is large, a better approximation of
$f$ is obtained, but with a larger Lipschitz constant for the gradient of $f$
(i.e., less smoothness).
Consequently, we approximate $w_*$ in \eqref{eq:min_max_opt} with
\begin{align}
w_*^\beta=\argmin_w f^\beta(w)\,.
\label{eq:w_*^beta}
\end{align}

To perform gradient-based convex optimization on $f^\beta$, we calculate its
gradient via
\begin{align}
    \label{eq:softmax_grad}
\nabla_w f^\beta(w) =  \frac{1}{n}\sum_i \mathbb{E}_{Y}(\nabla_w f_i(Y_i, w))\,,
\end{align}
where $Y$ is a random variable on $\mathcal{Y}$ with its probability distribution
function being the Boltzmann distribution of a system with the configuration set
$\mathcal{Y}$, energy function $-f_i(y, w)$, and inverse temperature $\beta$:
\begin{align}
\mathbb{P}(Y=y) =
\frac{\exp(\beta f_i(y, w))}{\sum_{y'\in \mathcal{Y}}\exp(\beta f_i(y, w))},
\quad y\in \mathcal{Y}\,.
\end{align}

\subsection{Quantum Gibbs Sampling}
\label{sec:q-gibbs}

We now describe the above problem in terms of Hermitian matrices we intend to
simulate on a quantum computer. For each $i$, we assume that the range of
$f_i(-, w): \mathcal{Y} \to \mathbb R$ corresponds (with an opposite sign)
to the spectrum of a diagonal Hermitian matrix $H_i^w$. We assume we have access
to oracles for $H_i^w$ and its partial derivatives. The oracles act on two
registers via
\begin{align}
\ket{k}\ket{z} \mapsto
\ket{k}\ket{z \oplus (H_i^w)_{kk}}
\end{align}
and
\begin{align}
\ket{k}\ket{z} \mapsto
\ket{k}\ket{z \oplus (\partial_j H_i^w)_{kk}}\quad (\forall j),
\end{align}
where $k$ ranges over the elements of $\mathcal Y$ and $z$ is any computational
basis state.

Here and in what follows, $\partial_j$ is used as an
abbreviation of the notation of partial
derivatives with respect to the vector $w$, that is,
$\partial_j= \partial/\partial w_j$.
The assumption is that access to such an oracle would require logarithmically
many qubits in the size of the Hermitian matrix
and the output precision of the oracle. For instance, if $f_i(-, w)$ is a
quadratic polynomial in binary variables with quadratic and linear coefficients
dependent on $w$, as is the case in many machine learning tasks
(refer to \cref{sec:intro} and in particular \cref{ex:Ising_model}),
we may associate the energies of an Ising model with logarithmically many
spins to the function $f_i(-, w)$.

The operator $\max^\beta$ would then simply be the negative free energy of $H_i^w$:
\begin{align}
\label{eq:max-beta-f-i}
{\max_{y \in \mathcal{Y}}}^\beta f_i(y,w)
= \frac{1}{\beta}\log \tr (\exp(-\beta H_i^w))\,.
\end{align}
Applying stochastic gradient descent for minimizing \eqref{eq:min_max_func} would require
calculation of the gradients of $f_i(y, w)$ with respect to $w$, which is
equal to
$\tr(A\rho)$ where $\rho= \frac{\exp(-\beta H)}{\tr(\exp(-\beta H))}$ is the
Gibbs state's density matrix and $A$ is the observable associated with the partial
derivatives
\begin{align}
\label{eq:partials-max-beta-f-i}
\partial_k {\max_y}^\beta f_i(y,w) =
\tr\left[ \left(- \partial_k H_i^w\right) \rho\right]\,.
\end{align}

This is exactly the type of quantity studied in \cite{van2017quantum}. They show
that for $N \times N$ diagonal matrices $H$ and $A$, such that $\|A\|\leq 1$
(in the operator norm)
and given an inverse temperature $\beta$, the quantity $\tr(A\rho)$ can be approximated
up to an additive error of at most $\theta$ with high probability.
We need to slightly modify the result of \cite{van2017quantum} for our application
and for reference we first state their result.
\begin{prop}[Corollary~12 in \cite{van2017quantum}]
Let $A, H \in \mathbb R^{n\times n}$
be diagonal matrices with $\|A \| \leq 1$. An additive $\theta$-approximation of
$\tr(A \rho)$ can be computed using $O(\sqrt n / \theta)$ queries to $A$ and $H$,
and $\widetilde O(\sqrt n / \theta)$ other gates\footnote{In the rest of this
paper, such a characterization of the number of quantum gates will be referred
to as being ``{almost of the same order}''.}.
\end{prop}

For our application, we need to include the contribution of the operator norms
of $A$ and $H$ in the complexity. We also require control over the success
probability of the approximation obtained in the above statement, which can be
boosted using the powering lemma for fully polynomial randomized approximation
schemes \cite{jerrum1986random}. Given these considerations, the complexity of
quantum Gibbs sampling from classical functions is as follows.

\begin{prop}[Quantum Gibbs sampling]
\label{prop:diagonal_gibbs}
Let $A, H \in \mathbb R^{n\times n}$
be diagonal matrices with $\|A\| \leq \Delta$ and $\|H\| \leq K$, and
$\rho$ be the Gibbs state of $H$ at inverse temperature $\beta$.
An additive $\theta$-approximation of
$\tr(A \rho)$ can be computed with a success probability of at least $1 - \zeta$
using $O(\frac{\sqrt n \Delta\beta K}{\theta} \log \frac{1}{\zeta}) $ queries to
$A$ and $H$, with the number of other quantum gates being almost of the same
order.
\end{prop}

We now impose boundedness conditions on the functions $f_i$ in
\eqref{eq:max-beta-f-i} before applying the quantum Gibbs sampling algorithm to
compute the partial derivatives \eqref{eq:partials-max-beta-f-i}.

\begin{condition}
\label{cond:oracle}
Let $\mathcal Y$ be a finite set and
$f: \mathcal{Y} \times \mathbb R^D \to \mathbb R$ be a real-valued function.
We assume that
(1) there exist $\Delta > 0$ such that
$\|\partial_k f\| \leq \Delta$ for all $w \in \mathbb R^D$,
$y \in \mathcal Y$, and $k= 1, \ldots, D$; and,
(2) there exist quantum oracles acting on
$O(\mathrm{polylog} (\frac{1}\delta, |\mathcal Y|))$
qubits to compute $f$ and $\partial_k f$ with an additive error of $\delta$.
\end{condition}

We can now derive the computational complexity of using quantum Gibbs sampling
to estimate the partial derivatives in \eqref{eq:partials-max-beta-f-i}.

\begin{thm}
\label{thm:quantum-derivatives}
Let $f_i: \mathcal{Y} \times \mathbb R^D \to \mathbb R$ be a real-valued
function satisfying \cref{cond:oracle}. Then the gradients of
\eqref{eq:max-beta-f-i} with respect to the parameter vector $w$ can be
calculated in
$O\left(\frac{D\sqrt{|\mathcal{Y}|} \Delta\beta F}{\theta}
\log \frac{D}{\zeta }\right)$
queries to the oracles of $f_i$ with the number of other gates being almost of
the same order. Here $F$ is a bound on values of all $f_i$, and $1-\zeta$ is the
probability that all dimensions of the gradient estimate have an additive error
of at most $\theta$.
\end{thm}

\subsection{A-SAGA: Approximate SAGA}
\label{sec:noisy-saga}

Stochastic average gradient (SAG) \cite{schmidt2017minimizing} and its variant
SAGA \cite{defazio2014saga} are two optimization methods that are specifically
designed for minimizing the sum of finitely many smooth functions.
SAG and SAGA usually perform better than standard stochastic gradient
descent (SGD) \cite{robbins1985stochastic}. The general
idea behind SAG and SAGA is to store the gradients for each of the $n$ functions
in a cache, and use their summation to obtain an estimation of the full gradient.
Whenever we evaluate the gradient for one (or some) of the functions, we update
the cache with the new gradients. Although the gradients in the cache are for
previously visited points, if the step size is small enough,
and the functions are smooth, the gradients in the
cache will not be too far from the current gradients; thus, using them
will reduce the error of estimation of the full gradient, leading to an
improved convergence rate.

Our approach is to use SAGA to minimize the smooth, strongly
convex objective function $f^\beta(w)$ to approximate the minimum of the original
nonsmooth objective function $f$. Quantum Gibbs sampler will provide approximations of the
derivatives of the functions $\max^\beta_y f_i (y, w)$ (but not exactly), as
stated in \cref{thm:quantum-derivatives}. Consequently, we need
to revisit the convergence of SAGA in the presence of errors in
calculating the gradients and do so in the following sections. In this section,
the notation $\langle - , - \rangle$ is used to represent the inner products
of real numbers.

\begin{condition}
\label{cond:convex}
Each function $f_i$ is $\mu$-strongly convex,
resulting in each $g_i(w)=\max_y f_i(y,w)$ being
$\mu$-strongly convex. The vector $w$ is restricted to a convex set
$\mathcal{W}$. Furthermore, the gradients of $f_i(y,w)$ are $\ell$-Lipschitz
smooth, and the partial derivatives are bounded by
\begin{align}
    \label{eq:min_max_grad_bound_1}
    \Delta = \max_{w,i,j,y} \left\| \partial_j [f_i(y,w)] \right\|,
\end{align}
where the maximum ranges over every index $i$, every $y \in \mathcal Y$,
every $w\in \mathcal{W}$, and every $j$-th component of $w$.
\end{condition}

We note that \cref{cond:convex} has important differences with \cref{cond:rsg_1}.
In \cref{cond:convex}, the functions $f$ have Lipschitz continuous gradients,
whereas in \cref{cond:rsg_1} there was not such restriction. Also, in
\cref{cond:convex}, we impose a bound $\Delta$ on the partial derivatives,
whereas in \cref{cond:rsg_1}, $M$ is a bound on the subgradients.

In the approximate SAGA algorithm (A-SAGA) presented in \cref{alg:A-SAGA},
we have an estimate of the gradient with an additive
error of at most $\theta/3$
\footnote{The division by $3$ was chosen to simplify the formulae.}
in each partial derivative appearing in the gradient.
Here the update rule for SAGA from \cite[Equation (1)]{defazio2014saga} has been
modified to take into account an approximation
error $ \Theta^{t+1}$ in step $t+1$, where the vector $\Theta^{t+1}$ comprises
all the additive errors, (that arise from the Gibbs sampler
in the following section\footnote{In fact, the Gibbs sampler is used to calculate each
directional derivative up to an additive error. Therefore, the approximation errors in
all the terms in the square brackets in line \ref{line:main-update} of \cref{alg:A-SAGA} contribute to
the bound on $\Theta$. More precisely, if the Gibbs sampler calculates the
derivatives with error $\frac{\theta}3$,
then $\|\Theta^{t+1}\|  \leq \theta$.}), that is,
\begin{align}
    \label{eq:main-update-error}
    \Theta^{t+1} = \Upsilon^{t+1}_j - \Upsilon^{t}_j +
    \frac{1}{n}\sum_{i=1}^n \Upsilon^t_i.
\end{align}

\begin{alg*}
\noindent\framebox{\begin{minipage}[t]{0.986\textwidth}\small

\begin{algorithmic}[1]
    \Procedure{A-SAGA}{$w^0\in \mathcal{W}$: initial point,
                       $T \in \mathbb{N}$: number of iterations,
                       $g_1, \ldots, g_n$: functions}
    \State $\widetilde{g_1}', \dots, \widetilde{g_n}':= 0$
    \Comment{Initialize the cached values of the gradients to zero.}

    \State $\widetilde{G}':= 0$
    \Comment{Initialize the summation of the cached gradients to zero.}

    \For{$t \in \{0, \dots, T-1\}$}
    \State $j \sim \text{Unif}\{1, \dots, n\}$
    \Comment{Draw the index $j$ uniformly at random.}
    \State $\widehat{g}_j':= g'_j(w^{t})+\Upsilon^{t+1}_j$
    \Comment{Estimate the gradient up to an additive error $\Upsilon^{t+1}_j$.}
    \State $\temp^{t+1}:=
    w^t-\st\left[ \widehat{g}_j'- \widetilde{g_j}'
    + \frac{1}n\widetilde{G}' \right]$ \label{line:main-update}
    \Comment{Take a descent step according to the SAGA formula.}
    \State $w^{t+1}:= \Pi_{\mathcal{W}}(\temp^{t+1})$
    \Comment{Project the point onto the set $\mathcal{W}$.}
    \State $\widetilde{G}':= \widetilde{G}' + \widehat{g}_j'- \widetilde{g_j}' $
    \Comment{Update the summation of cached gradients with the new gradient.}
    \State $\widetilde{g_j}':= \widehat{g}_j' $
    \Comment{Update the cached gradient with the new gradient.}
    \EndFor
    \State \Return ${w}^T$
    \EndProcedure
\end{algorithmic}
\end{minipage}}
\caption{Pseudocode for the A-SAGA algorithm.}
\label{alg:A-SAGA}
\end{alg*}

Note that for all vectors $\Upsilon^t_i$, every element has an absolute value
of at most $\theta/3$. Based on the definition of $\Theta^{t+1}$
from \eqref{eq:main-update-error}, we can conclude that every element of the
vector $\Theta^{t+1}$ is at most $\theta$.

Following the approach of \cite{defazio2014saga} we find a bound for
$\|w^t - w_*\|$ using the Lyapunov function $\mathbb{T}$ defined as
\begin{align}
\label{eq:asaga_lyaponov_func}
\begin{split}
\mathbb{T}^t
&:= \mathbb{T}(w^t, \{\phi_i^t\}_{i=1}^n)\\
&:= \frac{1}n \sum_i g_i(\phi^t_i)-f(w_*)
- \frac{1}n \sum_i\left\langle g'_i(w_*),\phi^t_i-w_*\right\rangle
+c\left\|w^t-w_*\right\|^2,
\end{split}
\end{align}
by proving the inequality $\mathbb{E}[\mathbb{T}^{t+1}]\leq(1-\frac{1}{\tau})\mathbb{T}^t$.
The next theorem proves a similar inequality in the case that an additive error
on the gradients exists.

\begin{thm}
\label{thm:strong-convex-thm}
Let the precision of a subroutine calculating the gradients of $g_i$ at every
point be
\begin{align}
\label{eq:theta-setting}
\theta = \min\left\{ \frac{1}{\sqrt{D}}, \frac{\mu \|w^t - w_* \|^2}
{2\sqrt{D}\left( \frac{3}{34 L}+2\|w^t - w_* \| \right)}\right\}\,.
\end{align}
Then there exists a choice of step sizes $\gamma$ in line \ref{line:main-update} of \cref{alg:A-SAGA},
$c$ in \eqref{eq:asaga_lyaponov_func}, and $\tau > 0$ such that for all $t$,
$\mathbb{E}[\mathbb{T}^{t+1}]\leq(1-\frac{1}{\tau})\mathbb{T}^t\,.$
\end{thm}

\begin{rem}
As shown in the proof of this theorem in the appendix (refer to
\eqref{eq:asaga_params}), the step size $\gamma$ does not depend on the
strong convexity parameter $\mu$. This is a desirable property called
``adaptivity to strong convexity''.
\end{rem}

The next theorem provides the time complexity of optimizing the smooth
approximation $f^\beta$ via A-SAGA, depending on the condition number $L/\mu$,
where $L$ is the Lipschitz constant of the gradient of $f^\beta$.
We let $w_*^\beta$ denote the minimizer of the smooth function $f^\beta$.

\begin{thm}
\label{thm:asaga_complexity_1}
Under \cref{cond:convex}, and given $\epsilon$ as a target precision, A-SAGA
finds a point $w$ such that
$\mathbb{E}\left[ \|w-w_*^\beta\|^2 \right] \leq \epsilon$ using
$O\left( \left({n}{} + \frac{\beta D\Delta^2 + \ell}{ \mu}\right)
\left(\log \frac{n}{\epsilon(\beta D\Delta^2 + \ell)}\right)\right)$
iterations.
\end{thm}

\begin{rem}
The number of gradient evaluations in \cref{thm:asaga_complexity_1} is
$O\left(\log \frac{1}{\epsilon}\right)$
in terms of $\epsilon$ only.
Also, based on \eqref{eq:asaga_params}, we have $\theta=O(\epsilon)$.
\end{rem}

\subsection{Using A-SAGA to Optimize the Nonsmooth Objective Function}
\label{sec:complexity-pred}

In this section, we analyze the
inverse temperature $\beta$ at which sampling from the quantum Gibbs sampler has to
happen in order for $w_*^\beta$ to be a sufficiently good approximation of the
original optimum $w_*$.

\begin{lem}
\label{lem:min_max_error_1}
To solve the original problem \eqref{eq:min_max_opt} with
$\epsilon$-approximation, it suffices to optimize the smooth approximation
\eqref{eq:smooth_min_max_func} for $\beta > \frac{\log |\mathcal Y|}{\epsilon}$
with precision $\epsilon - \frac{\log |\mathcal Y|}{\beta}$.
\end{lem}

\begin{lem}
\label{lem:min_max_error_2}
In solving problem \eqref{eq:w_*^beta} with A-SAGA we have
\begin{align}
\mathbb{E}\left[ f^\beta(w^t) - f^\beta(w_*) \right]
\leq \frac{L}{2} C_0 \left(1-\frac{1}{\tau}\right)^t.
\end{align}
\end{lem}

The above two lemmas are useful for achieving an approximation of the optimal
\emph{value} of $f$ by doing so for $f^\beta$.

\begin{thm}
\label{thm:a-saga-converge-in-value}
Under \cref{cond:convex}, A-SAGA applied to the function $f^\beta$ at
$\beta = \frac{2\log |\mathcal{Y}|}{\epsilon}$ requires
$O\left(\left(\frac{D \Delta^2\log |\mathcal{Y}|}{ \mu\epsilon}
+ \frac{\ell}{ \mu}\right)
\left(\log \frac{n}{\epsilon} \right)\right)$
iterations to find a point $w$ at which the original function value $f$ is
$\epsilon$-close to its minimum in expectation, that is,
$\mathbb{E}\left[ f(w)-f(w_*) \right] \leq \epsilon$,
provided $\epsilon$ is sufficiently small.
\end{thm}

\begin{rem}
The number of gradient evaluations in \cref{thm:a-saga-converge-in-value} is
$O\left(\frac{1}{\epsilon}\log \frac{1}{\epsilon}\right)$ in terms of $\epsilon$.
We note that in terms of the precision factor, the optimal scaling
for optimizing \eqref{eq:min_max_func} is
$O(\frac{1}\epsilon)$ \cite{shamir2013stochastic,nesterov2005smooth}, matching
the theoretical optimal bound. Our result is close to optimal (up to a
logarithmic factor).

It is also interesting to observe that based on
\eqref{eq:asaga_params}, we have $\theta = O(\sqrt \epsilon)$, which
means that to optimize $f$, we do not need as much precision as
for optimizing $f^\beta$. Surprisingly, the error in gradient evaluations could
be orders of magnitude larger than the desired precision and the algorithm
would still converge with the same rate as in SAGA.
\end{rem}

Finally, it is easy to use the previous theorem and the definition of strong
convexity to show convergence of A-SAGA to an approximation of the optimal
solution of $f$.

\begin{cor}
\label{thm:a-saga-converge-in-solution}
With the same conditions as \cref{thm:a-saga-converge-in-value}, A-SAGA finds
a point $w$ at which $\mathbb{E}\left[ w-w_* \right] \leq \epsilon$,
provided $\epsilon$ is sufficiently small, using
$O\left(\left(\frac{D\Delta^2\log |\mathcal{Y}|}{ \mu\epsilon}
+ \frac{\ell}{ \mu}\right)
\left(\log \frac{n}{\mu\epsilon}\right)\right)$
iterations.
\end{cor}

\begin{ex}
A special case of practical importance is when the functions $f_i$ are a linear
function in $w$ plus an $L^2$ regularizer. In this case our objective function
to minimize is
\begin{align}
f(w) = \lambda\frac{\|w\|^2}{2} +
\frac{1}{n} \sum_{i=1}^n \max_{y\in \mathcal Y} \{ a_{i,y}w+b_{i,y} \}\,.
\end{align}

Let $\mathcal{W}= \mathbb B^D (0, 1)$ be the unit ball centred at the origin
of $\mathbb R^D$, where $D$ is the dimension of $w$. For the linear functions,
the Lipschitz constant of the gradients is $0$, as
the gradient does not change. For the regularizer $\lambda\frac{\|w\|^2}{2}$,
the Lipschitz constant of the gradient is $\lambda$.
Therefore, $\ell = \lambda$.
For the bound on the partial derivatives of the
functions, we have $\Delta = \lambda + \max_i \max_j \max_y |a_{i,y, j}|$,
where $a_{i,y, j}$ is the $j$-th element of the vector $a_{i,y}$.

A further special case is when the functions $f_i$ minus the regularizer remain
linear in $w$ but are quadratic in $y$, e.g., the energy function of
an Ising model (refer also to \cref{ex:Ising_model})
\begin{align}
f(w) = \lambda\frac{\|w\|^2}{2} +
\frac{1}{n} \sum_{i=1}^n \max_{y\in \mathcal Y} \{ y J_i y^{T} + h_i y^{T} \}\,,
\end{align}
where $\mathcal{Y} = \{-1, 1 \}^m$, $J_i \in \mathbb{R}^{m\times m}$,
and $h_i\in \mathbb{R}^m$,
for an Ising model with $m$ particles.
Here the vector $w$ includes all the elements of the matrices $J_i$ and vectors
$h_i$ for all $i$.
In this case $\mathcal{W}$ is the unit ball of dimension $D= nm(m+1)$ around
the origin. Similar to the previous example
we still have $\ell = \lambda$. For the bound on the gradient of the
functions, we have $\Delta=\lambda + 1$, where we use the fact
that the elements of $y$ are in $\{-1, 1\}$.
\end{ex}

\subsection{Comparison of SAGA and A-SAGA}

We now compare our previous result to the case wherein exact gradients are
available. That is, we optimize $f^\beta$ using SAGA, and use it to approximate
the optimal solution of $f$.

\begin{thm}
\label{thm:saga-complexity-beta-and-original}
Under \cref{cond:convex}, and given $\epsilon$ as a target precision, SAGA uses
\begin{align}
\label{eq:saga-f-beta}
O\left( \left(n + \frac{ \beta D \Delta^2 +\ell}{\mu}\right)
\left(\log \frac{n}{\epsilon (\mu n + \beta D \Delta^2 + \ell)}
\right)\right)
\end{align}
gradient evaluations to find a point in the $\epsilon$-neighbourhood of
$w_*^\beta$ defined in \eqref{eq:w_*^beta} and
\begin{align}
\label{eq:saga-f-itself}
O\left(\left(n + \frac{ D\Delta^2 \log|\mathcal{Y}|}{\epsilon\mu}
+ \frac{\ell}{\mu}\right) \left(\log \frac{n}{\epsilon}\right)\right)
\end{align}
gradient evaluations to find an $\epsilon$-approximation of the optimal value
of $f$.
\end{thm}

It is clear that the scaling in \eqref{eq:saga-f-beta} with respect to all
parameters is similar to \cref{thm:asaga_complexity_1} and the scaling in
\eqref{eq:saga-f-itself} is similar to \cref{thm:a-saga-converge-in-value},
except for an extra $n$ term added in the first parentheses.

\begin{rem} We summarize the results of
\cref{thm:asaga_complexity_1,thm:a-saga-converge-in-value,thm:saga-complexity-beta-and-original} by observing that with
$O(\epsilon)$ and $O(\sqrt \epsilon)$ additive errors in gradient evaluations
in A-SAGA algorithm, its scaling for optimizing $f^\beta$ and $f$ remains similar to
SAGA, which does not assume any errors in gradient evaluations.
\end{rem}

\subsection{Q-SAGA: A Quantum Algorithm for Optimizing the Smooth Approximation}
\label{sec:q-saga}

In \cref{thm:asaga_complexity_1,thm:a-saga-converge-in-value}, we have assumed
that the additive error in calculating the partial derivatives is always at most
$\theta/3$. Using the quantum Gibbs sampler from \cref{sec:q-gibbs}, we can
guarantee such an upper bound only with a non-zero probability of failure.
As shown in \cref{thm:quantum-derivatives}, the gradients of the function
$\max^{\beta}_{y} f_i(y,w)$ can be estimated with additive
errors of at most $\theta$ in all partial derivatives appearing in the gradient
with a hight probability. We now propose a quantum algorithm, called Q-SAGA,
for optimizing the smooth approximation function $f^\beta(w)$
(by combining \cref{thm:quantum-derivatives,thm:asaga_complexity_1})
and for optimizing the original function $f$
(by combining \cref{thm:quantum-derivatives,thm:a-saga-converge-in-value}),
using a quantum Gibbs sampler.
Here $\beta$ is a fixed inverse temperature. The higher this value is, the more
accurate the approximation of $f(w)$ via $f^\beta (w)$ will be.

\begin{lem}
\label{lem:qsaga_grad}
\sloppy Under \cref{cond:oracle,cond:convex}, each gradient evaluation takes
$O\left(\frac{D^{1.5} \beta F \Delta}{\mu\epsilon}
\left(\frac{1}{\beta D \Delta^2 + \ell}+ \sqrt{\epsilon}\right)
\sqrt{|\mathcal{Y}|} \log(|\mathcal{Y}|)
\log \frac{D}{\zeta}\right)$
queries to the oracle for one of the $f_i$ with
the number of other quantum gates being almost of the same order,
where $1-\zeta$ is the probability of the Gibbs sampler returning
a gradient estimate whose additive errors in all partial derivatives are
at most $\theta$ as in \eqref{eq:theta-setting}.
\end{lem}

\begin{thm}
\label{thm:qsaga}
Under \cref{cond:oracle,cond:convex},
given sufficiently small $\epsilon > 0$ as a target precision, Q-SAGA finds a point
$w$ satisfying
$\mathbb{E}\left[ \|w-w_*^\beta\|^2 \right] \leq \epsilon$,
with a probability of at least $3/4$, in
$O\Bigg(
\frac{n\, D^{1.5} \beta F \Delta}{(\beta D \Delta^2 + \ell) \mu \epsilon}
\sqrt{|\mathcal{Y}|} \log(|\mathcal{Y}|)
\left(\log \frac{n}{\epsilon}\right)
\log\left(D\, n\, \log \frac{n}{\epsilon}\right)
\Bigg)$
queries to the oracle for one of the $f_i$ with the number of other quantum
gates being almost of the same order, when $f^\beta$ is sufficiently smooth
(i.e., the condition number $L/\mu$ is sufficiently small), and otherwise, in
$O\Bigg(
\frac{D^{1.5} \beta F \Delta}{\mu^2 \epsilon}
\sqrt{|\mathcal{Y}|} \log(|\mathcal{Y}|)
\left(\log \frac{n}{\epsilon}\right)
\log\left(D\, n\, \log \frac{n}{\epsilon}\right)
\Bigg)$
queries to the oracle for one of the $f_i$ with the number of other quantum
gates being almost of the same order. In both cases, the complexity is
$\widetilde O(\frac{1}{\epsilon})$ in terms of $\epsilon$ only.
\end{thm}

\begin{thm}
\label{thm:qsaga-original}
Under \cref{cond:oracle,cond:convex},
given sufficiently small $\epsilon > 0$ as a target precision,
Q-SAGA finds a point $w$ such that
$\mathbb{E}\left[ \|w-w_*\|^2 \right] \leq \epsilon$,
with a probability of at least $3/4$ in
$O\Bigg(
\left( \frac{D^{2.5} \Delta^{3} F\sqrt{|\mathcal{Y}|}
(\log^2 |\mathcal{Y}|)(\log D)}{\mu^2\epsilon^{1.5}} \right)
\left(\log \frac{n}{\epsilon} \right)
\log \left (\frac{D\Delta^2\log |\mathcal{Y}|}{ \mu\epsilon}
\left(\log \frac{n}{\epsilon} \right) \right)
\Bigg)$
queries to the oracle for one of the $f_i$ with the number of other quantum gates being
almost of the same order.
This is $\widetilde O(\frac{1}{\epsilon^{1.5}})$ in terms of $\epsilon$ only.
\end{thm}

\section{Numerical Experiments}
\label{sec:numerical}

\subsection{Synthetic Benchmark}
\label{sec:synthetic_exps}
We compare the optimization of the function $f$, as defined in
\eqref{eq:min_max_func}, with its smooth approximation $f^\beta$, as defined in
\eqref{eq:smooth_min_max_func}. To exclude the effects of sampler errors and
noise, we restrict our experiments to small instances (i.e., we restrict the
size of the sets $\mathcal{Y}$) in order to be able to find the value of
the softmax operator and its gradient exactly.
Also, for simplicity, we restrict our experiments to the case
where each $f_i$ is a linear
function of $w$ plus an $L^{2}$ regularizer:
\begin{align}
\label{eq:plane}
f_i(y, w) = \frac{\lambda}{2} \| w\|^2 + a_{i,y}^T (w-b'_i) + b_{i,y}
\qquad y\in \mathcal{Y}, \lambda \in \mathbb{R}^+, w\in \mathbb{R}^D\,.
\end{align}
Adding of the $L^2$ regularizer guarantees the strong convexity of $f$.

The elements $y\in \mathcal{Y}$
are used as indices for their corresponding $a_{i, y}$ and $b_{i, y}$ vectors.
All coefficient vectors $a_{i, y}$ and $b_{i, y}$ are randomly
generated according to the Cauchy distribution, and all vectors $b'_i$
are randomly generated according to a uniform distribution.
The reason we choose the Cauchy distribution for $a_{i, y}$ and $b_{i, y}$,
is its thick tail, which results in having occasional extreme
values for the coefficients. The reason we choose the uniform
distribution for $b'_i$ as opposed to normal or Cauchy distributions
is to avoid the functions $f_i$ having a similar
minimum, which makes the problem easy.

We generate a random objective function with $D=10$
parameters, that is, $w\in \mathbb{R}^{10}$, where $w$ is initialized to the vector
$w=(10, 10, \dots, 10)^T$. We use $200$ summand functions $f_i$, that is, $n=200$.
We set $\lambda=2$ and $\mathcal{Y} = \{1,2,\dots,  100\}$. We generate the
vectors $b'_i$ from the uniform distribution over the set $[0, 10000]^{10}$.

We benchmark four gradient descent schemes: (1) stochastic gradient descent (SGD)
applied to the smooth approximation $f^\beta$;
(2) SGD applied to the original nonsmooth function $f$ ({SubSGD}); (3) stochastic
subgradient descent with polynomial-decay averaging (SGDP)
\cite{shamir2013stochastic} applied to the original nonsmooth function $f$; and
(4) SAGA \cite{defazio2014saga} applied to the smooth approximation
$f^\beta$.

All methods have two tunable hyperparameters in common: (1) $\gamma_0$, the
initial step size gradient descent or its
variations; and (2) $c_\gamma\,$, a constant indicative of a schedule on $\gamma$
through the assignment of $\gamma_t= \frac{\gamma_0}{1+t c_\gamma}\, $ at
iteration $t$.
SGD and SAGA are applied to the smooth approximation $f^\beta$ and, as such, the
inverse temperature $\beta$ is a tunable hyperparmeter in these methods. In
contrast, {SubSGD} and SGDP are applied to the original nonsmooth objective
function. SGDP also has an additional hyperparmeter $\eta$, which is used to
define the polynomial-decay averaging scheme.
For each algorithm, we tune the hyperparameters via a grid search with respect
to a quantity we call \emph{hyperparameter utility} that is explained below.
We use the following values to form a grid in each case:
\begin{align}
  \beta &\in \{10^{-7}, 10^{-5}, \dots, 10^{0}\};\\
  \gamma_0 &\in \{10^{-7}, 10^{-5}, \dots, 10^{0}\};\\
  c_\gamma &\in \{0\} \cup \{10^{-4}, 10^{-3}, \dots, 10^{2}\};\, \text{and}\\
  \eta &\in \{1,2, \dots, 7\}.
\end{align}

We run each algorithm $20$ times with different seeds for random number
generation in the algorithm, which randomizes the choice of
index $i$ at each iteration of the algorithm, wherein
we perform $1000$ iterations, and track the progress on the original
nonsmooth objective function $f$. We emphasize that
the objective function remains the same over the $20$ runs and the seeds
are not used to regenerate the random objective function.

For each of the gradient descent schemes mentioned above,
and each hyperparameter setting, we calculate the average
objective value over all $20$ runs and all $1000$ iterations. For each
algorithm we calculate the following quantities:
(1)~total descent---the difference between the initial objective value and the
best value found over all 20 trials;
(2)~absolute ascent---the sum of the values of all ascents between any two consecutive iterations over all iterations of all 20 trials; and
(3)~hyperparameter utility---the absolute ascent divided by the total descent.

For each algorithm, we choose the hyperparameter setting that minimizes the
average objective value over $20$ runs and $1000$ iterations subject to the
constraint that its hyperparameter utility is less than $0.01$. We use this
constraint to avoid unstable hyperparameter settings. For instance, a very large
step size might reduce the objective value very quickly in the beginning but
fail to converge to a good solution.

\begin{table}
\begin{center}
\begin{tabular}{ |c|c|c|c|c|}
\hline
Algorithm & $\beta$ & $\gamma_0$ & $c_\gamma$ & $\eta$
\\ \hline
Smooth SGD & $10^{-4}$ & $10^{-2}$ & $10^{1}$ & N/A
\\ \hline
{Nonsmooth SGD} & N/A & $10^{-2}$ & $10^{1}$ & N/A
\\ \hline
SGDP & N/A & $10^{-3}$ & $0$ & $5$
\\ \hline
SAGA & $10^{-4}$ & $10^{-3}$ & $0$ & N/A
\\ \hline
$\beta$-10-SAGA & $10^{-7}-10^{-6}$ & $10^{-3}$ & $0$ & N/A
\\ \hline
\end{tabular}
\caption{The tuned hyperparameter values.}
\label{tab:min_max_fig_hyperparameters}
\end{center}
\end{table}

The value of the hyperparameters found by the grid search for each algorithm is
reported in \cref{tab:min_max_fig_hyperparameters}. Other than the four
methods discussed above, a final row called $\beta$-10-SAGA has been included,
a description of which follows.

We can see that for SAGA, we have $c_\gamma=0$, resulting in a constant step
size consistent with the theoretical proof of convergence of SAGA. For SGD and
{SubSGD}, we obtain $c_\gamma = 10$, which is also consistent with the
theoretical step sizes of $\sfrac{1}{\mu t}$ and $\sfrac{\eta}{\mu (t+\eta)}$,
respectively \cite{shamir2013stochastic}. Note that by the contribution of the
regularizer $r(w)=\frac{\lambda \| w\|^2}{2}$, we have $\mu \geq \lambda = 1$.
For SGDP, we see that the polynomial-decay averaging manages to work with
a constant step size, whereas to prove its theoretical convergences, a step
size of $\sfrac{\eta}{\mu (t+\eta)}$ is used.

\begin{figure}[b]
\centering
\includegraphics[width=0.5\textwidth]{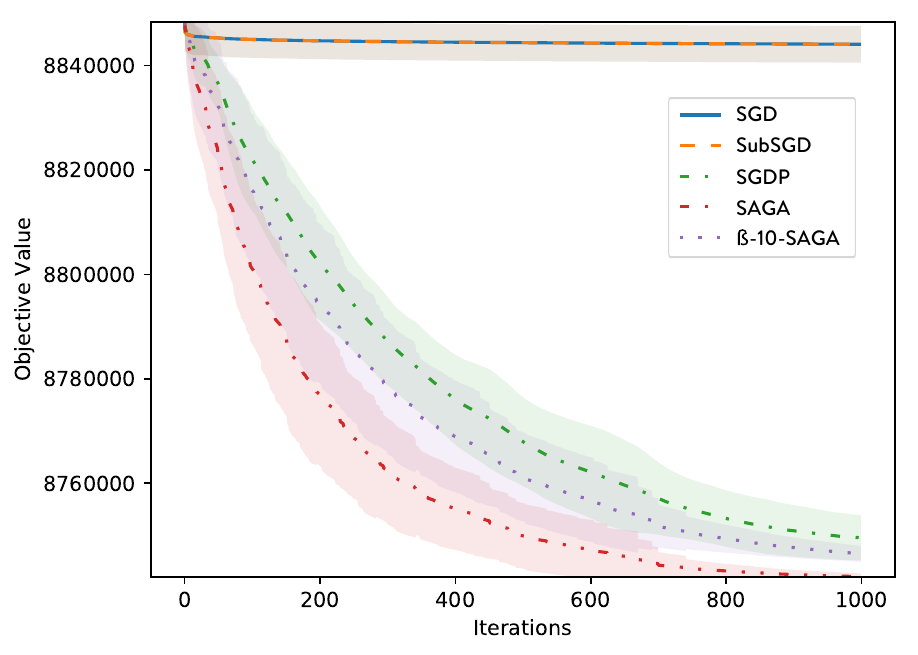}
\caption{The average objective value of five algorithms, SGD, {SubSGD}, SGDP,
SAGA, and $\beta$-10-SAGA, tracked as a function of gradient descent
iterations.}
\label{fig:min_max_fig}
\end{figure}

The five gradient descent schemes, SGD, {SubSGD}, SGDP, SAGA, and
$\beta$-10-SAGA, are compared in \cref{fig:min_max_fig}. The objective function
value is recorded over $20$ trials of each gradient descent scheme. For each
gradient descent iteration, the average objective value is shown using dashed
and dotted lines, alongside the standard deviation, shown using shaded regions.

We see that SGD and {SubSGD} perform poorly (at least for stable choices of
hyperparameters, e.g., their having small step sizes). SGDP results in a great
improvement, yet SAGA further outperforms it.
This is despite the fact that SAGA optimizes the original nonsmooth function
in $O(\frac{1}{\epsilon}\log \frac{1}{\epsilon})$ once applied to the smooth
approximation $f^\beta$, whereas SGDP converges theoretically in the provably
optimal rate of $O(\frac{1}{\epsilon})$.

We observe that the objective function value is around $8\times 10^6$.
Hence $\frac{1}{n}\sum \max f_i \approx \max f_i \approx 8\times 10^6$. Therefore,
at $\beta= 10^{-4}$, we have $\beta \max f_i \approx 800$.
In this regime, the Boltzmann distribution
from which we need to sample is very close to the delta function concentrated on
the (possibly degenerate) ground states.

In an alternative SAGA experiment, called $\beta$-10-SAGA, we have $\beta$ start
from $10^{-7}$ and in every $10$ iterations increase it by $10^{-8}$, resulting
in a final value of $1.1\times 10^{-6}$. As shown in \cref{fig:min_max_fig},
$\beta$-10-SAGA performs slightly worse than SAGA, although it is still better
than SGDP. However, $\beta \max f_i$ starts from around $0.8$ and approaches $8$
in the end, which is a more suitable temperature regime for a quantum Gibbs
sampler.

\subsection{Image Tagging as a Structured Prediction Task}
\label{sec:image-tagging}

In this section we consider the computer vision task of image tagging
to demonstrate an application of the smooth min-max objective functions
\eqref{eq:smooth_min_max_func}.
In image tagging, given an image we would like to assign all
the relevant tags (possibly more than one)
from a predefined set of tags {(e.g., ``cat'', ``dog'', ``river'', ``nature'')}.

We formulate this problem as a structured prediction problem and train a model
on a dataset of pairs of images and tags. To achieve this, we choose three
objective functions to benchmark:
\begin{enumerate}
\item The smoothed structural support vector machine (S3VM):
\begin{align}
f_{\text{S3VM}} (w;\beta) = \frac{1}{2} \lambda \| w\|^2
+ \sum_{(x,{y})\in \mathcal{S}}
{\max}^\beta_{y'} \left\{ \Delta(y', {y}) + s(x, y'; w) - s(x, y; w)\right\}.
\label{eq:s3vm-obj}
\end{align}
This objective function is the smooth approximation of
\eqref{eq:ssvm_unconstrained} using the softmax operator \eqref{eq:softmax}.
Here, $s$ is a scoring function, as explained in
\cref{subsec:structured-prediction}.

\item The conditional log-likelihood (CL) objective function
\begin{align}
f_{\text{CL}}(w; \beta) = \frac{1}{2} \lambda \| w\|^2
+ \sum_{(x,y)\in \mathcal{S}} {\max}^{\beta}_{y'}
\left\{ s(x, y'; w) - s(x, y; w) \right\} \,.
\label{eq:cl-obj}
\end{align}
Similar to S3VM, this objective function is in the form of the smooth min-max
objective function \eqref{eq:smooth_min_max_func} but does not incorporate the
loss function $\Delta$.

\item The Jensen risk bound (JRB) objective function
\begin{align}
f_{\text{JRB}}(w; \beta) &= \frac{1}{2} \lambda \| w\|^2
+ \sum_{(x,y)\in \mathcal{S}}
\log \mathbb{E}_{Y_B} \left( \beta \exp({\Delta}(Y_B,y) )\right)\,.
\label{eq:jrb-obj}
\end{align}
This objective function incorporates the scoring function $s$ and the loss
function $\Delta$ in a different way. Here $Y_B$ is a random variable
that follows a Boltzmann distribution with energy $-s(x, y; w)$ at inverse
temperature $\beta$. More details about the derivation of the CL and JRB
objective functions can be found in \cite{gimpel2010softmax1}.
\end{enumerate}

Recall the notation used in \cref{subsec:ssvm}. In our image tagging task, let
$\mathcal{X}$ be the set of all possible images,
and $\mathcal{Y}= \{-1, 1\}^\ell$ be the set of
all possible labels. The labels are $\ell$-dimensional binary vectors, with
each component representative of presence or absence of a tag in the image.
We would like to find the feature function $\Phi(x, y; w_0)$ with parameter
$w_0$. Let $\Phi_0:\mathcal{X}\times\mathcal{W}_0\rightarrow \mathbb{R}^{\eta}$
be a feature function, where the first argument from $\mathcal{X}$ is an image,
the second argument from $\mathcal{W}_0$ is a parameter, and the output is a
real vector with $\eta\in\mathbb{N}$ dimensions. The function $\Phi_0(x; w_0)$
serves as a base feature function in the construction of $\Phi(x, y; w_0)$.
The function $\Phi_0(x; w_0)$ can be any function. In our experiments, we use
a convolutional neural network (CNN) as a feature extractor for this purpose,
with $w_0$ denoting its weights.

One way to define $\Phi$ based on $\Phi_0$ is as follows: we design $\Phi_0$
(i.e., the CNN) such that the dimension of its output is identical to the size of the
labels: $\eta = \ell$. Let ``triu'' denote the vectorized upper triangle of
its square matrix argument. We then define
\begin{align}
\label{eq:Phi}
\Phi(x,y; w_0) =
\begin{pmatrix}
\text{triu}(y y^T)\\
\Phi_0(x; w_0) \circ y\\
y
\end{pmatrix}
,
\end{align}
where $\circ$ is the element-wise product. Note that $\Phi_0(x; w_0) \circ y$
is well-defined because $\eta=\ell$ and the two vectors $\Phi_0(x; w_0)$ and $y$
have identical dimensions.
The result is $\Phi(x, y; w_0) \in \mathbb R^d$ for some $d \in \mathbb N$. Let
$w \in \mathbb R^d$ be the parameter vector of our structured prediction model.
We then define the scoring function $s$ as
\begin{align}
\label{eq:energy}
s(x, y; w) &= w^T \Phi(x, y; w_0)=
\begin{pmatrix} \theta_1^T & \theta_2^T & \theta_3^T
\end{pmatrix} \Phi(x, y; w_0)
\\&= \theta_1^T \text{triu}(y y^T)
+ \theta_2^T \left[\Phi_0(x; w_0) \circ y\right]
+ \theta_3^T y\,. \nonumber
\end{align}

One can then interpret $\theta_1$ as a control parameter on the relationship
between pairs of labels $y_i$ and $y_j$. The parameter vector $\theta_2$
controls the effect of the features extracted from the CNN. The parameter vector
$\theta_3$ controls the bias of the values of $y_i$, as some tags are less likely
to be present and some are more likely. Note that the formula $s(x, y; w)$
in \eqref{eq:energy} is quadratic in $y$.

We choose the function $\Delta$ to be the Hamming distance
for two reasons. Firstly, the error in the predictions made in image tagging is also
calculated using the Hamming distance between the true label and the predicted
label. Secondly, the Hamming distance is a linear function of $y'$, and
therefore $\Delta(y', y) + s(x, y'; w)$ remains quadratic in
$y'$. This simplifies the gradient calculations of both $f_{\text{S3VM}}$
and $f_{\text{JRB}}$ to expectations with respect to the Boltzmann
distribution of an Ising model rather than more complicated distributions.

\subsubsection{Numerical Results}

\begin{figure*}
\centering
\includegraphics[width=0.6\textwidth]{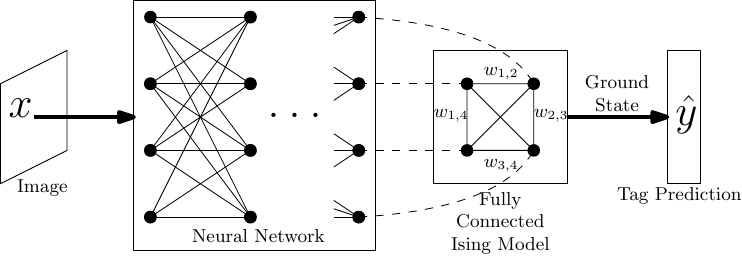}
\caption{Image tagging architecture. The image $x$ is fed to
a neural network to extract features. The features then are passed to
an Ising model the ground state of which determines the prediction.}
\label{fig:architecture}
\end{figure*}

We use the MIRFLICKR dataset \cite{huiskes2008mir}, which consists of 25,000
images and 38 tags. We randomly select 20,000 images for the training set, 2500
images for the validation set, and the remaining 2500 images for the test set.
This dataset consists of an extended tag set with more than 1000 words. For
this demonstration, we restrict our numerical experiments to the smaller set of
38 tags to ensure fast convergence of Monte Carlo simulations. However, we note
that the extended set of labels in this dataset is well within reach of Monte
Carlo simulations using today's high-performance computing platforms.

We train a pre-trained AlexNet \cite{krizhevsky2012imagenet}, a convolutional
neural network, on the training data, to predict the tags.
This is done on an Ubuntu machine with 32 AMD Ryzen CPU cores, 128~GB of memory,
and an Nvidia Titan~V GPU. We train AlexNet
using the binary cross entropy objective function between its output layer and
the true labels.
We call this model a \emph{baseline} in what follows.
We fix the baseline, which acts as a feature extractor,
and feed its output features to an Ising model
which acts as a predictor. We then train the weights of
the Ising model with three different objective functions, namely $f_{\text{CL}}$,
$f_{\text{S3VM}}$, and $f_{\text{JRB}}$. This is inspired by
\cite{chen2015learning}, wherein the output of an AlexNet network is fed to
an Ising model in a very similar fashion, and trained using the $f_{\text{CL}}$
objective function.
The architecture of the model is shown in \cref{fig:architecture}.

In the training mode, we use the standard stochastic gradient descent algorithm, with a parameter
$\lambda$ adjusting the $L^2$ regularizer of $\lambda\| w\|^2/2$ that is added to
the objective functions, and a parameter $\gamma$ as the step size, which is
kept constant during the training. We consider four training epochs, where, in
each epoch, we go through each data point of the training data exactly once, in a random order. In
this experiment, we use single-spin flip Gibbs sampling at a constant inverse
temperature $\beta$ as our sampling subroutine to compute a Monte Carlo
estimation of the objective function's gradient. Due to our choice of using only a
subset of tags to train and test over, our Ising model instances consist of 38
variables and a fully connected architecture. For each instance, we perform 200
sweeps and collect 200 samples.
In total, we have three hyperparameters, namely $\gamma$,  $\lambda$, and
$\beta$. We tune the hyperparameters by performing a grid
search over the values
\begin{align}
\gamma&= \{ 10^{-8},10^{-7}, 10^{-6}, 10^{-5}\},\\
\lambda&= \{0.0, 10^{-6}, 10^{-4}, 10^{-2}\},\, \text{and}\\
\beta&=  \{3^{-1}, 3^{0}, 3^{1}, 3^{2}\}\,.
\end{align}

A last architecture considered is that of an extension of the baseline with a
fully connected feedforward layer with sigmoid activations. This model has been added
in order to compare the extensions of the baseline with undirected architectures
(e.g., the Ising model) versus a feedforward layer using a similar number of
parameters. The Ising model has a fully connected graph with $\binom{38}{2}+38=741$
parameters and we use a fully connected feedforward layer with 38
nodes, which amounts to $38^2+38=1482$ parameters.
We use the Adam algorithm for optimization \cite{kingma2014adam}
implemented in the PyTorch library \cite{paszke2017automatic} with 300 epochs.
We tune the step size parameter $\gamma$ using a gridsearch over the values
\begin{align}
\gamma= \{ 10^{-5},10^{-4}, 10^{-3}, 10^{-2}, 10^{-1}\},
\end{align}
while all other hyperparameters of Adam are left at their default values
($\beta_1=0.9, \beta_2=0.999$).

\begin{table*}[t]
\begin{center}
\begin{tabular}{ |c|c|c||c|c|c||c|}
 \hline
 Model  & Validation Error & Test Error & $\gamma$ & $\lambda$ & $\beta$ & $\beta_{\text{eff}}$
  \\   \hline
baseline &   2.6844 &  2.7052  & N/A & N/A & N/A & N/A
\\  \hline
baseline + S3VM  & \textbf{2.6568}  & \textbf{2.6900}   & $10^{-7}$ & 0.0  & $3^{1}$ & $[60.3720,133.0482]$
\\  \hline
baseline + CL &   2.6696 & 2.6996 &  $10^{-6}$ & $10^{-6}$ & $3^{1}$ & $[53.0406,118.1979]$
\\  \hline
baseline + JRB &   2.6580 & 2.6956 &  $10^{-7}$ & $10^{-6}$ & $3^{1}$ & $[55.4559,122.7675]$
\\  \hline
baseline + FC &   2.7236 & 2.7656 & $10^{-2}$ & 0.0 & N/A & N/A
\\  \hline
\end{tabular}
\caption{Image tagging results.}
\label{tab:result_image_tagging}
\end{center}
\end{table*}

In \cref{tab:result_image_tagging}, we summarize the performance of the various
methods and values of tuned hyperparameters. The baseline architecture is that
of AlexNet. The three subsequent lines report the performance of extensions of
the baseline with an Ising model trained using objective functions
\eqref{eq:s3vm-obj} to \eqref{eq:jrb-obj}. The last row is an extension of the
baseline with a single feedforward fully connected layer with sigmoid
activations and the binary cross entropy objective. The performance of each
method is measured via the average Hamming distance between the predicted
labels and the true labels in terms of the number of bits. Therefore,
the validation and test error values reported in the second and third
columns are the average numbers of wrong tags (including the missing tags and the
incorrect additional tags).

We observe that all three extensions of the baseline
with an Ising model improve the baseline validation errors with the S3VM
objective function, resulting in the
greatest improvement of $\sim0.03$ tags. The same
observation holds for the test errors, although the improvement is smaller in
this case, at $\sim0.015$ for S3VM. In contrast, the last row of the table shows
that the extension of AlexNet with an additional trailing fully connected layer
not only increases the training and test errors, but it increases the difference
of these two quantities from 0.0208 to 0.042, which hints to a slightly worsened
generalization. It remains to be practically verified whether slight
improvements observed in this table would become more significant for larger
image tagging tasks.

We observe that the values of $\lambda$ in all cases are either $0$ or very
small. However, this might be an artifact of having small numbers of
parameters in our model ($\binom{38}{2}+38 = 741$), making the model immune to
over-fitting.

In the final column of \cref{tab:result_image_tagging}, we report the range of
the effective thermodynamic $\beta$ denoted by $\beta_{\text{eff}}$ for each
method. The effective $\beta$ is the product of the nominal value $\beta$ and the
absolute value of the ground state energy of the Ising model over different
images. The interval reported in this table is the range of $\beta_\text{eff}$
over the images in the test set.

\begin{figure*}[t]
\centering
\setlength{\fboxsep}{0pt}%
\setlength{\fboxrule}{0.2pt}
\begin{subfigure}[t]{0.332\textwidth}
  \centering
  \fbox{\includegraphics[width=0.89\textwidth]{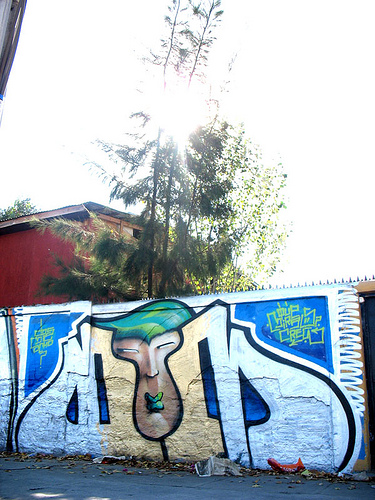}}
  \caption{Test image 7520}
  \label{fig:im7520}
\end{subfigure}%
\begin{subfigure}[t]{0.3205\textwidth}
  \centering
  \fbox{\includegraphics[width=1.05\textwidth]{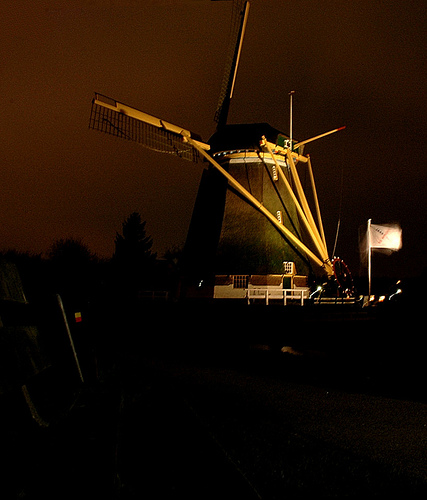}}
  \caption{Test image 10177}
  \label{fig:im10177}
\end{subfigure}
\begin{subfigure}[t]{0.339\textwidth}
  \centering
  \fbox{\includegraphics[width=0.86\textwidth]{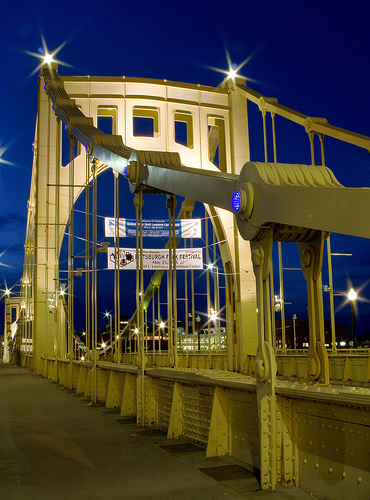}}
  \caption{Test image 21851}
  \label{fig:im21851}
\end{subfigure}
\resizebox{\textwidth}{!}{
\begin{tabular}{l|l|l|l}
  image number & 7520 & 10177 & 21851 \\
  \midrule
  true labels &
  plant\_life,\!~sky,\!~structures,\!~tree &
  night,\!~sky,\!~structures,\!~transport &
  night,\!~sky,\!~structures
  \\
  baseline &
  people,\!~plant\_life,\!~sky,\!~structures,\!~tree &
  night,\!~plant\_life,\!~sky,\!~structures,\!~sunset,\!~transport,\!~tree &
  indoor,\!~male,\!~people,\!~structures
  \\
  baseline\!\!~+\!\!~S3VM &
  plant\_life,\!~sky,\!~structures,\!~tree &
  night,\!~plant\_life,\!~sky,\!~structures,\!~sunset,\!~tree &
  indoor,\!~male,\!~people,\!~structures
  \\
  baseline\!\!~+\!\!~CL &
  people,\!~plant\_life,\!~sky,\!~structures,\!~tree &
  night,\!~plant\_life,\!~sky,\!~structures,\!~sunset,\!~transport,\!~tree &
  male,\!~people,\!~structures
  \\
  baseline\!\!~+\!\!~JRB &
  plant\_life,\!~sky,\!~structures,\!~tree &
  night,\!~plant\_life,\!~sky,\!~structures,\!~sunset,\!~tree &
  indoor,\!~male,\!~people,\!~structures
  \\
  baseline\!\!~+\!\!~FC &
  male,\!~people,\!~plant\_life,\!~sky,\!~structures,\!~tree &
  night,\!~plant\_life,\!~sky,\!~structures,\!~sunset,\!~transport,\!~tree &
  male,\!~people,\!~sky,\!~structures
\end{tabular}}

\caption{Sample tags generated by the different models.
In \Cref{fig:im7520}, \Cref{fig:im10177}, and \Cref{fig:im21851},
we see that S3VM has respectively
decreased, increased, and did not affect
the error,
compared to the baseline.}
\label{fig:image_tagging}
\end{figure*}

In \cref{fig:image_tagging}, we see three examples from the test set.
Finally, we wish to remark that we would have needed to solve much larger
problems and perform many more sweeps of Monte Carlo simulations had we used the
complete set of tags. The fully connected architecture is not imposed by the
problem we are solving. The use of much sparser connectivity graphs could result
in viable feature extractors as well. These are future areas of development that
can be explored using quantum computing and classical high-performance computing
platforms.

\section{Conclusion}
\label{sec:conclusion}

In this paper, we introduced quantum algorithms for solving the min-max
optimization problem that appears in machine learning applications.
We first studied the variant A-SGDP of the subgradient descent with
polynomial-decay averaging (SGDP) \cite{shamir2013stochastic} which takes
into account incorrect calculations of the subgradients as long as they are
bounded. This allows the quantum minimum finding algorithm of
\cite{durr1996quantum} to find the subgradients used in subgradient method
efficiently while leaving room for a small probability of failure.
The combination of A-SGDP and the quantum minimum finding algorithm results in the
quantum algorithm Q-SGDP for solving the original (nonsmooth) min-max
optimization problem without resorting to smooth approximations.
Despite an almost-linear scaling in terms of the precision factor $\epsilon$,
this quantum algorithm provides a speedup in terms of the size of the discrete
optimization space $\mathcal Y$. We showed that Q-SGDP solves the original
min-max problem in $\widetilde O(\frac{1}{\epsilon}\sqrt{|\mathcal Y|}\log
\frac{1}{G})$, in which the value $G$ is the minimum gap we encounter during the
runtime of the algorithm which is unknown beforehand.

Secondly, we studied a
variant of SAGA (which we call A-SAGA) that takes into account an additive error on the
calculation of gradients. This has allowed us to use a quantum Gibbs sampler as
a subroutine of A-SAGA to provide estimations of the gradients and optimize the
smooth approximation of the min-max problem. We called the conjunction of A-SAGA
with the quantum Gibbs sampler Q-SAGA. We have shown that A-SAGA can give an
approximation of the solution of the smooth approximation of the original min-max problem in
$O(\log \frac{1}{\epsilon})$ gradient evaluations, provided the additive error
is in $O(\epsilon)$. This scaling is, in fact, optimal
\cite{defazio2014saga,schmidt2017minimizing}.
We then used A-SAGA to solve the
original min-max problem in $\widetilde{O}(\frac{1}{\epsilon})$
gradient evaluations. We remark
that the best algorithms \cite{shamir2013stochastic,nesterov2005smooth}
for solving the original min-max problem use $O(\frac{1}\epsilon)$
gradient evaluations. This is the case if the gradients are calculated exactly.
We conclude that in the presence of additive errors in estimating the
gradients, our result is close to optimal.

Consequently, the quantum algorithm Q-SAGA solves the smooth approximation
of the original min-max problem in $\widetilde{O}(\frac{1}\epsilon)$
queries to the associated quantum oracles.
We also analyzed the usage of Q-SAGA, not to solve the smooth prediction problem,
but to approximate a solution to the original min-max problem. In order to do
this, the temperature has to be assigned proportional to $\epsilon$. In total, this
results in $\widetilde O(\frac{1}{\epsilon^{1.5}} \sqrt{|\mathcal Y|})$
queries to the associated oracles. Despite a slightly worse scaling in precision,
this algorithm does not have the dependence on the minimum gap $G$ as in Q-SGDP.

Finally, we have provided results from several numerical experiments. In particular,
we compared the performance of SGD in two cases: with all sampling subroutines
performed at a constant temperature, and with the temperature decreasing across
iterations according to a schedule. We observed that the scheduled temperature
slightly improves the performance of SGD. We believe that studying the temperature
schedule would be an interesting avenue of research. In particular, it would be
beneficial to gain
an understanding of the best practices in scheduling temperature during
SGD. It would also be interesting to provide a theoretical analysis of the
effect of the temperature schedule in SGD.
As we have seen in our experiments, using a temperature schedule seems not to be
consistent with SAGA since the cache of old gradients then comes from other
temperatures. Another avenue of future research would be to adapt or modify SAGA
so as to overcome this caveat.

Our successful image tagging experiments used only 38 English words as candidate
tags. The MIRFLICKR dataset provides a thousand English words as candidate tags,
but conducting
an experiment of this size was not feasible with the computational resources
available to us. Our goal is to pursue efficient Gibbs sampling approaches in
quantum and high-performance computation in order to achieve similar results
in larger image tagging tasks.
In fact, our work proposes a general approach for quantum machine learning using
a quantum Gibbs sampler. In this approach, the network architecture consists of a
leading directed neural network serving as a feature extractor, and a
trailing undirected neural network responsible for smooth prediction based on
the feature vectors.

\section{Acknowledgement}

The authors thank Mark~Schmidt, who motivated our initiation of
this project and provided technical feedback throughout. We further thank
Ronald~de~Wolf, Matthias~Troyer, Joran~van~Apeldoorn, Andr\'as~Gily\'en,
Austin~Roberts, and Reza~Babanezhad for useful technical discussions, and
Marko~Bucyk for helpful comments and for reviewing and editing the mxanuscript.
This project was fully funded by 1QBit. P.~R. further acknowledges the support
of the government of Ontario and Innovation, Science and Economic Development
Canada.

\bibliography{main}

\section{Appendix}
\label{sec:appendix}

\subsection{Proof of \cref{thm:SGDP}}

Because the failure probability in solving the $\argmax_y f_i(y,w^t)$
in each iteration is
$\zeta = 1/(4T)$, the probability of not having any failure in $T$ iteration
is at least $3/4$ as we can verify
\begin{align}
\prod_{t=1}^T(1-\zeta )
= (1-\frac{1}{4T})^T \geq 1 - T\frac{1}{4T} = 1 - \frac{1}{4} = \frac{3}{4}.
\end{align}
Conditioned on not seeing any failure, we can directly use
\cite[Theorem 4]{shamir2013stochastic}
to conclude that with the stepsize $\gamma^t = \frac{1}{\mu t}$ we need
$T=O(\eta^{1.5}\frac{M}{\mu \epsilon})$ to satisfy
$\mathbb{E}\left[ |f(w^T)-f(w_*)| \right] \leq \frac{\epsilon}{4}$.
Using Markov inequality, we can conclude that
\begin{align}
\mathbb{P}\Big[|f(w^T)-f(w_*)| \leq \epsilon
~\Big|~ \text{\small No failure in $T$ iterations}
\Big] \geq \frac{3}{4}.
\end{align}
Hence the probability that $|f(w^T)-f(w_*)| \leq \epsilon$ is higher than
$1/2$ as we can verify
\begin{align}
\begin{split}
\mathbb{P}\left[ |f(w^T)-f(w_*)| \leq \epsilon \right]
& = \mathbb{P}\Big[|f(w^T)-f(w_*)| \leq \epsilon ~\Big|~
\text{\small No failure in $T$ iterations} \Big] \\
& \times \mathbb{P}\Big[\text{\small No failure in $T$ iterations}\Big]
\geq \left(\frac{3}{4}\right)^2 \geq \frac{1}{2}.
\end{split}
\end{align}
Finally hiding the dependence of complexity on $\eta$, we get the result.

\subsection{Proof of \cref{thm:qSGDP}}

By multiplying the number of iterations from \cref{thm:SGDP}
by the query complexity found in \cref{lem:grover_max}, we obtain a scaling of
$O\left(\frac{ \eta^{1.5} M}{\mu \epsilon}
\sqrt{|\mathcal{Y}|} \log (F/G) \log(1/\zeta) \right)$,
where $F$ is a bound on the absolute values of $f_i$.
Using the value of $\zeta$,
the result  follows.

\subsection{Proof of \cref{prop:diagonal_gibbs}}

This proposition is proven similarly to \cite[Corollary 12]{van2017quantum},
except that before applying \cite[Lemma 9]{van2017quantum}, the success
probability has to be boosted. We restate \cite[Lemma 9]{van2017quantum}.

\begin{lem}
\label{lem:probability_amplitude_estimation}
Suppose we have a unitary $U$ acting on $q$ qubits such that
$U\ket{0\ldots0} = \ket{0}\ket{\psi} + \ket{\Phi}$, with $\bra{0}\otimes I
\ket{\Phi}= 0$ and $\|\psi\|^2 = p \geq p_{\min}$ for some known bound $p_{\min}$.
Let $\mu \in (0, 1]$ be the allowed multiplicative error in our estimation of
$p$. Then, with $O\left(\frac{1}{\mu \sqrt{p_{\min}}} \right)$ uses of $U$ and
$U^{-1}$, and using $O\left(\frac{q}{\mu \sqrt{p_{\min}}} \right)$ gates on the
$q$ qubits, we obtain a $\widetilde p$ such that $|p - \widetilde p| \leq \mu p$
with a probability of at least $4/5$.
\end{lem}

We note that this lemma provides a multiplicative approximation of $p$ and, as
such, it is a fully polynomial randomized approximation scheme as defined in
\cite{jerrum1986random}. Therefore, for any target rate $\zeta$, the powering
lemma \cite[Lemma 6.1]{jerrum1986random} asserts that with $O(\log(1/\zeta))$
repetitions of the algorithm and choosing the median of the returned values as
the approximation $\tilde p$, its success probability can be boosted to
$1 - \zeta$.

\subsection{Proof of \cref{thm:quantum-derivatives}}

With $H_w^i$ diagonal real-valued matrices realizing $f_i(-, w)$ and
$A= \partial H_w^i$, the boundedness of derivatives, $\|f'_i(w)\|$ for all $w$,
is equivalent to $\|A\| \leq \Delta$.
In order to estimate all partial derivatives in the gradient with an
additive error of at most $\theta$ successfully with a probability of at least
$1 - \zeta$,
we may calculate each of the partial derivatives with a success probability of
at least $1 - \zeta/D$, because
$(1-\frac{\zeta}{D})^D \geq 1- \frac{\zeta}{D}D = 1-\zeta\,.$
By the previous corollary, each partial derivative is therefore calculated in
$O\left(\frac{\sqrt{|\mathcal{Y}|}\Delta\beta F}{\theta} \log\frac{D}{\zeta}\right)$
and, since there are $D$ such partial derivatives, the result follows.

\subsection{Proof of \cref{thm:strong-convex-thm}}

Defazio~et~al. prove three lemmas in \cite{defazio2014saga}. Following their
convention, all expectations are taken with respect to the choice of $j$ at
iteration $ t+1 $ and conditioned on $w^t$ and each $g'_i(\phi_i^t)$
and additive errors $\Upsilon^t_j$, unless otherwise stated.

\begin{lem}
\label{lem:ip-bound}
Let $f(w)= \frac{1}n \sum_{i=1}^n g_i(w)$. Suppose each $g_i$ is $\mu$-strongly
convex and has Lipschitz continuous gradients with  constant $L$. Then for all
$w$ and $w_*$:
\begin{align}
\begin{split}
\left\langle f'(w),w_*-w\right\rangle
& \leq \frac{L-\mu}{L}\left[f(w_*)-f(w)\right]
-\frac{\mu}{2}\left\|w_*-w\right\|^2 \\
& -\frac{1}{2Ln}\sum_i\left\|g'_i(w_*)-g'_i(w)\right\|^2
-\frac{\mu}{L}\left\langle f'(w_*),w-w_*\right\rangle.
\end{split}
\end{align}
\end{lem}

\begin{lem}
\label{lem:grad-diff-phii}
For all $\phi_i$ and $w_*$:
\begin{align}
\frac{1}n &\sum_i\left\|g'_i(\phi_i)-g'_i(w_*)\right\|^2
\leq 2L\left[ \frac{1}n \sum_i g_i(\phi_i)-f(w_*)
- \frac{1}n \sum_i\left\langle g'_i(w_*), \phi_i - w_*\right\rangle\right].
\end{align}
\end{lem}

The last lemma in \cite{defazio2014saga} is only true if the error in the A-SAGA
update rule is disregarded. We therefore restate this lemma as follows.
\begin{lem}
\label{lem:wchange}
For any $\phi_i^t$, $w_*$, $w^t$ and $\alpha>0$, with $\temp^{t+1}$ as defined in
SAGA, if
\begin{align}
X = g'_j(\phi_j^t)-g'_j(w^t) + f'(w_*)-\frac{1}{ n}\sum_i g'_i(\phi_i^t)\,,
\end{align}
it holds that
\small
\begin{align}
\mathbb{E}[X] &= f'(w^t) - f'(w_*),\quad \text{and}\\
\label{eq:X_bound_1}
\begin{split}
\mathbb{E}\|X\|^2 \leq &
(1+\alpha^{-1}) \mathbb{E}\left\| g'_j(\phi_j^t)-g'_j(w_*)\right\|^2
+ (1+\alpha) \mathbb{E}\left\| g'_j(w^t)-g'_j(w_*)\right\|^2
- \alpha \left\| f'(w^t)-f'(w_*) \right\|^2.
\end{split}
\end{align}
\normalsize
\end{lem}

We are now ready to state the proof of \cref{thm:strong-convex-thm}.

\begin{proof}[\cref{thm:strong-convex-thm}]
The first three terms in $\mathbb{T}^{t+1}$ can be bounded in a way similar to
the proof of \cite[Theorem 1]{defazio2014saga}:
\small
\begin{align}
\mathbb{E}\left[ \frac{1}n \sum_i g_i(\phi_i^{t+1})\right]
&=  \frac{1}n f(w^t)+\left(1- \frac{1}n \right)
 \frac{1}n \sum_i g_i(\phi_i^t),\quad \text{and}\\
\mathbb{E}\left[- \frac{1}n \sum_i\left\langle g'_i(w_*),
\phi_i^{t+1}-w_*\right\rangle \right]
&= - \frac{1}n \left\langle f'(w_*),w^t-w_*\right\rangle
- \left(1- \frac{1}n \right) \frac{1}n \sum_i\left\langle
g'_i(w_*),\phi_i^t-w_*\right\rangle.
\end{align}
\normalsize
The last term is bounded by the inequality
\begin{align}
c\left\|w^{t+1}-w_*\right\|^2
= c\left\|\Pi_{\mathcal{W}} (\temp^{t+1})-
\Pi_{\mathcal{W}} [w_*-\st f'(w_*)]\right\|^2
\leq c\left\|\temp^{t+1}- w_*+\st f'(w_*)\right\|^2\,,
\end{align}
by the optimality of $w_*$ and non-expansiveness of the projection operator $\Pi_{\mathcal{W}}$.
We can now bound the expected value of the right-hand side of this inequality
in terms of $X$ and $\|w^t - w_*\|$ by expanding the quadratics.
\begin{align}
\begin{split}
c\mathbb{E}\left\|\temp^{t+1}-w_*+ \st f'(w_*)\right\|^2\,
& =\, c\mathbb{E}\left\|w^t-w_*+ \st X + \st\Theta^{t+1}\right\|^2
\\
=\, c\left\|w^t-w_*\right\|^2
& + \left\{
2c\mathbb{E}\left[\left\langle \st X+ \st\Theta^{t+1}, w^t-w_*\right\rangle\right]
+c\mathbb{E}\left\|\st X + \st \Theta^{t+1} \right\|^2 \right\} \\
=\, c\left\|w^t-w_*\right\|^2
& + \!\begin{aligned}[t]
\Big\{
-2c\st\left\langle f'(w^t)-f'(w_*), w^t-w_*\right\rangle
+2c\st\mathbb E\left[\left\langle \Theta^{t+1}, w^t-w_*\right\rangle\right]\\
+c\st^2 \mathbb{E}\left\|X \right\|^2
+2 c\st^2 \mathbb E \left[\langle\Theta^{t+1}, X \rangle\right]
+c\st^2 \mathbb E\left\|\Theta^{t+1} \right\|^2 \Big\}
\end{aligned}
\end{split}
\end{align}
Using Jensen's inequality applied to the square root function,
in the second inequality below,
and then using $\sqrt{x}\leq \frac{1}{2}+\frac{x}{2}$,
we have
\begin{align}
\mathbb E \left[\langle\Theta^{t+1}, X \rangle\right] \leq
\theta \sqrt D \mathbb E\left[\|X\|\right] \leq
\theta \sqrt D \sqrt{\mathbb E\left[\|X\|^2\right]}
\leq
\frac{\theta \sqrt D }{2}+ \frac{ \theta \sqrt D \mathbb E \|X\|^2}{2}\,.
\end{align}
We now apply \cref{lem:wchange} and the assumption that
$\|\Theta^{t+1}\| \leq \theta\sqrt D$.
\begin{align}
\begin{split}
& \hspace{-5mm}
c\mathbb{E}\left\|\temp^{t+1}-w_*+ \st f'(w_*)\right\|^2 \,\\
\leq\, & c\left\|w^t-w_*\right\|^2
+
\!\begin{aligned}[t]
&\left\{\vphantom{\sum}
-2c\st\left\langle f'(w^t)-f'(w_*), w^t-w_*\right\rangle
+2c\st\mathbb E\left[\left\langle \Theta^{t+1}, w^t-w_*\right\rangle\right] \right.
\\ &\left.
+ \left(c\st^2 (1+\theta\sqrt D)\right) \mathbb{E}\left\|X \right\|^2
+ c\st^2 \theta \sqrt D
+c\st^2 \mathbb E\left\|\Theta^{t+1} \right\|^2
\vphantom{\sum}\right\}
\end{aligned}
\\
\leq\, &
c\left\|w^t-w_*\right\|^2 +
\!\begin{aligned}[t]
&\left\{-
2c\st\left\langle f'(w^t),w^t-w_*\right\rangle
+2c \st \left\langle f'(w_*),w^t-w_*\right\rangle
+2c\st \theta\sqrt D \|w^t-w_*\|
\right.\\
&- \left(c \st^2  (1 + \theta\sqrt D)\right)
 \alpha \left\|f'(w^t)-f'(w_*) \right\|^2\\
&+\left(1+\alpha^{-1}\right)
\left(c \st^2 (1 + \theta\sqrt D)\right)\mathbb{E}\left\|g'_j(\phi_j^t)
-g'_j(w_*)\right\|^2\\
&+\left(1+\alpha\right)\left(c \st^2  (1 + \theta\sqrt D)\right)
\mathbb{E}\left\|g'_j(w^t)
-g'_j(w_*)\right\|^2
\\
&\left. + c\st^2 \theta\sqrt D + c\st^2 \theta^2 D \right\}.
\end{aligned}
\end{split}
\end{align}
We now apply \cref{lem:ip-bound} and \cref{lem:grad-diff-phii} to respectively
bound $-2c\st\left\langle f'(w^t),w^t-w_*\right\rangle$
and $\mathbb{E}\left\|g'_j (\phi_j^t)-g'_j(w_*)\right\|^2$:
\small
\begin{align}
\begin{split}
c\mathbb{E}\left\|w^{t+1}-w_*\right\|^2
\leq \left(c- c\st\mu \right)\left\|w^t-w_*\right\|^2
+
\left\{ \vphantom{\sum^2_1}
\left((1 +  \theta \sqrt D)(1+\alpha)c\st^2-\frac{c\st }{L}\right)
\mathbb{E}\left\|g'_j(w^t)-g'_j(w_*)\right\|^2 \right.\\
-\frac{2c\st(L-\mu)}{L}\left[f(w^t)-f(w_*)
-\left\langle f'(w_*),w^t-w_*\right\rangle \right]
- c\st^2 ( 1 +  \theta \sqrt D)\alpha\left\|f'(w^t)-f'(w_*) \right\|^2\\
 +2\left(1 +  \theta \sqrt D\right)\left(1+\alpha^{-1}\right)c\st^2 L
\left[ \frac{1}n \sum_i g_i(\phi_i^t)-f(w_*)
- \frac{1}n \sum_i\left\langle g'_i(w_*),
\phi_i^t-w_*\right\rangle \right]\\
\left.
 + c\st^2 \theta\sqrt D + c\st^2 \theta^2 D + 2c \st \theta\sqrt D\|w^t-w_*\|
 \vphantom{\sum^2_1}\right\}.
\end{split}
\end{align}
\normalsize
As in \cite[Theorem 1]{defazio2014saga}, we pull out a $\frac{1}\tau$ factor of
$\mathbb{T}^t$ and use the above inequalities, taking into account the contributions from the
three error terms above:
\small
\begin{align}
\begin{split}
\mathbb{E}[\mathbb{T}^{t+1}]-&\mathbb{T}^t \leq -\frac{1}{\tau}\mathbb{T}^t +
\left( \frac{1}n -\frac{2c\st(L-\mu)}{L} -2c\st^2\mu\alpha (1 +  \theta \sqrt D) \right)
\left[f(w^t)-f(w_*)-\left\langle f'(w_*),w^t-w_*\right\rangle \right]\\
&+\left(\frac{1}{\tau}+2(1+\alpha^{-1})(1 + \theta \sqrt D )c\st^2L- \frac{1}n \right)
\left[ \frac{1}n \sum_i g_i(\phi_i^t)-f(w_*)- \frac{1}n \sum_i\left\langle
g'_i(w_*),\phi_i^t-w_*\right\rangle \right]\\
& +\left(\frac{1}{\tau}-\st\mu\right)c\left\|w^t-w_*\right\|^2
+\left((1+\alpha)\st(1 + \theta \sqrt D)-\frac{1}{L}\right)
c\st \mathbb{E}\left\|g'_j(w^t)-g'_j(w_*)\right\|^2\\
& +
 \left\{
 c\st^2 \theta\sqrt D + c\st^2 \theta^2 D + 2c \st \theta\sqrt D\|w^t-w_*\|
\right\}
\end{split}
\end{align}
\normalsize
According to a lemma that will follow (\cref{lem:saga_params}), we can ensure
that all round parentheses in the first three lines are non-positive by
setting the parameters according to
\small
\begin{align}
\label{eq:asaga_params}
\begin{split}
\gamma = \frac{1}{2(1+\alpha) L}\,,\,
c = \frac{4}{n\gamma}\,,\,
\alpha = 16 \,,\,
\frac{1}{\tau} = \min\left\{
\frac{1}{2n}, \frac{\gamma \mu}{2}\right\}\,,\,
\theta = \min\left\{ \frac{1}{\sqrt{D}}, \frac{\mu \|w^t - w_* \|^2}
{2\sqrt{D}\left( \frac{3}{34 L}+2\|w^t - w_* \| \right)}\right\}.
\end{split}
\end{align}
\normalsize
With this setting of the parameters,
\begin{align}
\left(\frac{1}{\tau}-\st\mu\right)c\left\|w^t-w_*\right\|^2
+ \left\{ c\st^2
\theta\sqrt D + c\st^2 \theta^2 D + 2c \st \theta\sqrt D\|w^t-w_*\|\right\}
\leq 0 \,.
\end{align}
Using the non-negativity of the expressions in square brackets completes the
proof.
\end{proof}

In the following lemma we derive an appropriate value for
the parameters that we used in
\cref{thm:strong-convex-thm} such that all the necessary
inequalities in the proof of \cref{thm:strong-convex-thm}
are satisfied.

\begin{lem}
\label{lem:saga_params}
Let $\delta = (1+\theta\sqrt{D})$. In order to satisfy all the inequalities
\begin{align}
\label{eq:saga_cond_1}
&\frac{1}{n} - 2c\gamma \left(\frac{L-\mu}{L}
+ \gamma \mu \alpha \delta\right) \leq 0\,,\\
\label{eq:saga_cond_2}
&\frac{1}{\tau} + 2\left(1+ \frac{1}{\alpha}\right) \delta c \gamma^2L
- \frac{1}{n} \leq 0\,,\\
\label{eq:saga_cond_3}
&\left(\frac{1}{\tau} - \gamma \mu\right) \|w^t - w_* \|^2 + 2\gamma^2\theta\sqrt{D}
+ \gamma^2\theta^2 D + 2\gamma \theta \sqrt{D} \|w^t - w_*\| \leq 0\,,
\\
\label{eq:saga_cond_4}
&\left(1+\alpha\right) \gamma \delta - \frac{1}{L} \leq 0\,,
\end{align}
it is sufficient to have
\small
\begin{align}
\gamma = \frac{1}{2(1+\alpha) L}\,,\,
c = \frac{4}{n\gamma}\,,\,
\alpha = 16 \,,\,
\frac{1}{\tau} = \min\left\{\frac{1}{2n}, \frac{ \gamma \mu}{2}\right\}\,,\,
\theta = \min\left\{ \frac{1}{\sqrt{D}}, \frac{\mu \|w^t - w_* \|^2}
{2\sqrt{D}\left( \frac{3}{34 L}+2\|w^t - w_* \| \right)}\right\} .
\end{align}
\normalsize
\end{lem}

\begin{proof}
In what follows, we enumerate the steps required to satisfy all
inequalities in the statement.
Having lower and upper bounds on the value of $\delta$ is useful
to this end. To impose an upper bound on $\delta$, we assume
\begin{align}
\label{eq:saga_theta_2}
\theta \leq \frac{1}{\sqrt{D}},
\end{align}
resulting in
\begin{align}
\label{eq:saga_delta_1}
1 \leq \delta \leq 2.
\end{align}
For \eqref{eq:saga_cond_4}, using the upper bound from \eqref{eq:saga_delta_1}
we have
\begin{align}
\left(1 + \alpha\right) \gamma \delta - \frac{1}{L}
\leq 2 \left(1 + \alpha\right) \gamma - \frac{1}{L}.
\end{align}
Hence we can satisfy \eqref{eq:saga_cond_4} by setting
\begin{align}
\label{eq:saga_gamma_1}
\boxed{\gamma = \frac{1}{2(1+\alpha) L}}\,\,.
\end{align}
For \eqref{eq:saga_cond_1}
we consider the two cases of $\frac{L}{\mu} > 2$ and $\frac{L}{\mu} \leq 2$.
When $\frac{L}{\mu} > 2$,
\begin{align}
\frac{1}{n} - 2c\gamma \left(\frac{L-\mu}{L}
+ \gamma \mu \alpha \delta\right) \leq
\frac{1}{n} - 2c\gamma \left(\frac{L-\mu}{L}\right)
< \frac{1}{n} - c\gamma\,.
\end{align}
It therefore suffices to have
\begin{align}
\label{eq:saga_c_1}
c \geq \frac{1}{n\gamma}.
\end{align}
Alternatively, if $\frac{L}{\mu} \leq 2$,
\begin{align}
\begin{split}
\frac{1}{n} - 2c\gamma \left(\frac{L-\mu}{L}
+ \gamma \mu \alpha \delta\right)
&\leq \frac{1}{n} - 2c\gamma \left(\gamma \mu \alpha \delta\right)
= \frac{1}{n} - 2c\gamma \left(\frac{1}{2(1+\alpha) L} \mu \alpha \delta\right)
\\
&\leq \frac{1}{n} - 2c\gamma \left(\frac{1}{2(1+\alpha) L} \mu \alpha \right)
= \frac{1}{n} - c\gamma \left(\frac{\alpha}{1+\alpha} \frac{\mu}{L}\right)\\
&\leq \frac{1}{n} - c\gamma \left(\frac{\alpha}{1+\alpha} \frac{1}{2}\right)
\leq \frac{1}{n} - \frac{c\gamma}{4}\,,
\end{split}
\end{align}
where in the last line we used $\frac{L}{\mu} \leq 2$ and
in the last inequality we made the assumption that
\begin{align}
\label{eq:saga_alpha_0}
\alpha \geq 1\,,
\end{align}
resulting in $\frac{\alpha}{1+\alpha} \geq \frac{1}{2}$. Consequently, to satisfy
\eqref{eq:saga_cond_1} it suffices to have
\begin{align}
\label{eq:saga_c_2}
c \geq \frac{4}{n\gamma}\,.
\end{align}
By combining \eqref{eq:saga_c_1} and \eqref{eq:saga_c_2}, we set
\begin{align}
\label{eq:saga_c_3}
\boxed{c = \frac{4}{n\gamma}}\,\,.
\end{align}
For \eqref{eq:saga_cond_2} we require that
\begin{align}
\label{eq:saga_cond_2_2}
2\left(1+ \frac{1}{\alpha}\right) \delta c \gamma^2L
- \frac{1}{n} < 0\,,
\end{align}
in which the inequality is strict (in order to assure $\frac{1}\tau$ is strictly
positive). Plugging in the values of $c$ from
\eqref{eq:saga_c_3} and $\gamma$ from \eqref{eq:saga_gamma_1},
and the upper bound on $\delta$ from \eqref{eq:saga_delta_1}, we have
\begin{align}
2 \left(\frac{1+\alpha}{\alpha}\right)
\delta \left(\frac{4}{n \gamma}\right) \gamma^2 L - \frac{1}{n}
\leq
4 \left(\frac{1+\alpha}{\alpha}\right)
 \left(\frac{4}{n \gamma}\right) \gamma^2 L - \frac{1}{n}
=\frac{8}{\alpha n} - \frac{1}{n}\,.
\end{align}
So, in order to satisfy \eqref{eq:saga_cond_2_2}, it suffices to have
$\frac{8}{\alpha n} - \frac{1}{n} < 0$, resulting in $\alpha > 8$. We may
therefore set
\begin{align}
\label{eq:saga_alpha_1}
\boxed{\alpha = 16}
\end{align}
in order to leave room for $\frac{1}{\tau}$ to be larger in the next step.
Note that this automatically satisfies \eqref{eq:saga_alpha_0}. With
this setting of $\alpha$, the left-hand side of \eqref{eq:saga_cond_2} is equal to
\begin{align}
\frac{1}{ \tau} + 2\left(1+ \frac{1}{\alpha}\right) \delta c \gamma^2L
- \frac{1}{n} = \frac{1}{ \tau} - \frac{1}{2n}\,.
\end{align}
To satisfy \eqref{eq:saga_cond_2}, it is sufficient to require that
\begin{align}
\label{eq:saga_tau_1}
\frac{1}{\tau} \leq \frac{1}{2n}\,.
\end{align}
For \eqref{eq:saga_cond_3} we need
\begin{align}
\frac{1}{ \tau} - \gamma \mu < 0\,,
\end{align}
where the inequality is strict. To satisfy this, we set
\begin{align}
\label{eq:saga_tau_2}
\frac{1}{\tau} \leq \frac{ \gamma \mu}{2}\,.
\end{align}
By combining \eqref{eq:saga_tau_1} and \eqref{eq:saga_tau_2}, we set
\begin{align}
\label{eq:saga_tau_3}
\boxed{\frac{1}{\tau}
= \min\left\{\frac{1}{2n}, \frac{ \gamma \mu}{2}\right\}}\,\,.
\end{align}
To satisfy
\eqref{eq:saga_cond_3}, using \eqref{eq:saga_tau_2}, we can instead satisfy
\begin{align}
\frac{- \gamma \mu}{2} \|w^t - w_* \|^2 + \gamma^2\theta\sqrt{D}
+ \gamma^2\theta^2 D + 2\gamma \theta \sqrt{D} \|w^t - w_*\| \leq 0\,.
\end{align}
Cancelling a $\gamma$ term and using the value of $\gamma$
from \eqref{eq:saga_gamma_1}, we would like to satisfy
\begin{align}
\label{eq:saga_theta_1}
\frac{\theta\sqrt{D}}{34L}
+\frac{\theta^2D}{34L}
+2\theta\sqrt{D}\|w^t - w_* \| \leq \frac{\mu}{2} \| w^t - w_*\|^2\,.
\end{align}
Using \eqref{eq:saga_theta_2}, we have
\begin{align}
\frac{\theta\sqrt{D}}{34L} + \frac{\theta^2D}{34L}
+ 2\theta\sqrt{D}\|w^t - w_* \| \leq
\frac{\theta\sqrt{D}}{34L} + \frac{2\theta \sqrt{D}}{34L}
+ 2\theta\sqrt{D}\|w^t - w_* \|
\end{align}
To satisfy \eqref{eq:saga_theta_1}, we may assume
\begin{align}
\theta \leq \frac{\mu \|w^t - w_* \|^2}{2\sqrt{D}
\left(\frac{3}{34 L}+2\|w^t - w_* \|\right)}\,,
\end{align}
and \eqref{eq:saga_theta_2}. Therefore, we set
\begin{align}
\boxed{\theta =
\min\left\{ \frac{1}{\sqrt{D}},
\frac{\mu \|w^t - w_* \|^2}{2\sqrt{D}
\left(\frac{3}{34 L}+2\|w^t - w_* \|\right)} \right\}}\,\,.
\end{align}
This completes the proof of the lemma.
\end{proof}

\subsection{Proof of \cref{thm:asaga_complexity_1}}

As in \cite[Corollary 1]{defazio2014saga}, we note that
$c\left\|w^t-w_*\right\|^2\leq \mathbb{T}^t$. Therefore, by chaining the
expectations
\begin{align}
\mathbb{E}\left[\left\|w^t-w_*\right\|^2\right]
\leq C_0 \left(1-\frac{1}\tau \right)^t,
\end{align}
where
\begin{align}
C_0 = \left\|w^0-w_*\right\|^2 + \frac{1}{c}
\left[f(w^0)-\left\langle f'(w_*),w^0-w_*\right\rangle-f(w_*)\right].
\end{align}
Therefore, we should have
\begin{align}
t \geq \frac{\log \frac{1}{\epsilon} + \log {C_0}}
{-\log\left( 1 - \frac{1}{\tau} \right)}\,.
\end{align}
Using the inequality $\log(1-x) \leq -x$, it suffices that
\begin{align}
t \geq \tau \left( \log \frac{1}{\epsilon} + \log {C_0} \right).
\end{align}
From \eqref{eq:asaga_params}, we know that
\begin{align}
\tau = \max\left\{2n, \frac{2}{\gamma\mu} \right\}
\leq \max\left\{2n, \frac{68L}{\mu} \right\},
\end{align}
where we have used the fact that $\theta \leq \frac{1}{\sqrt{D}}$.
So, we get
\begin{align}
t \geq \max\left\{{2n},\frac{68L}{\mu}\right\}
\left(\log \frac{1}{\epsilon} + \log {C_0}\right).
\end{align}

In \cite{beck2012smoothing} the authors prove that
${\max^\beta_{y \in \mathcal{Y}}} f_i(y, w)$
has Lipschitz continuous gradients with parameter $\beta D \Delta^2 + \ell$,
so the function $f^\beta$ has
Lipschitz continuous gradients with parameter $L = \beta D \Delta^2 + \ell$.
We also note that
$C_0 = O(1/c) = O(n/L) = O(\frac{n}{\beta D \Delta^2 + \ell})$.
Therefore, when $f^\beta$ is sufficiently smooth, that is,
\begin{align}
\frac{L}\mu = \frac{\beta D \Delta^2 + \ell}\mu \leq \frac{n}{34}\,,
\end{align}
we have
\begin{align}
t= O\left(n\left(\log\frac{1}{\epsilon}+
\log n - \log(\beta D \Delta^2 + \ell)\right)\right),
\end{align}
and otherwise
\begin{align}
t= O\left(\frac{\beta D \Delta^2 + \ell}{\mu}
\left(\log \frac{1}{\epsilon} +
\log n - \log (\beta D \Delta^2 + \ell)\right)\right).
\end{align}
We can combine these two bounds into one to complete the proof:
\begin{align}
t= O\left( \left({n}{} + \frac{\beta D \Delta^2 + \ell}{ \mu}\right)
\left(\log \frac{1}{\epsilon} +
\log n - \log (\beta D \Delta^2 + \ell)\right)\right).
\end{align}

\subsection{Proof of \cref{lem:min_max_error_1}}

The softmax operator ${\max}^\beta$ is an upper bound on the $\max$
function satisfying
\begin{align}
\label{eq:smooth_ineq}
\max_{y\in \mathcal{Y}}
\upsilon(y) \leq {\max_{y \in \mathcal{Y}}}^\beta \upsilon(y)
\leq \max_{y\in \mathcal{Y}}  \upsilon(y) + \frac{\log |\mathcal{Y}|}{\beta}\,,
\end{align}
for any function $\upsilon$ \cite{nielsen2016guaranteed}. Using this inequality
and the optimality of $w_*$ and $w_*^\beta$, it follows that
\begin{align}
f(w_*) \leq f(w_*^\beta) \leq
f^\beta(w_*^\beta) \leq f^\beta(w_*)
\leq f(w_*)+ \frac{\log |\mathcal{Y}|}{\beta}\,.
\end{align}
Therefore, $0 \leq f^\beta(w_*^\beta) - f(w_*) \leq \frac{\log |\mathcal Y|}{\beta}$.
So, in order to solve the original problem within an error of $\epsilon$, that is,
$f(w^t) - f(w_*) \leq \epsilon$,
it is sufficient to have $\frac{\log |\mathcal Y|}{\beta} < \epsilon$,
and
\begin{align}
f^\beta(w^t) - f^\beta(w_*^\beta) \leq \epsilon
- \frac{\log |\mathcal Y|}{\beta}\,.
\end{align}
Therefore
$f^\beta(w^t) - f(w_*) \leq \epsilon,$
and using the fact that $f(w^t)\leq f^\beta(w^t)$, we can conclude that
$f(w^t) - f(w_*) \leq \epsilon,$
completing the proof.

\subsection{Proof of \cref{lem:min_max_error_2}}

By the descent lemma \cite[Lemma 1.2.4]{nesterov2013introductory}, we have
\begin{align}
f^\beta(w) - f^\beta(w_*)
\leq \langle \nabla f^\beta(w_*), w - w_* \rangle+ \frac{L}{2}\|w-w_*\|^2.
\end{align}
The smoothness of the function $f^\beta$, the optimality of $w_*$, and the convexity
of $\mathcal{W}$ imply that
$\langle \nabla f^\beta(w_*), w - w_* \rangle \leq 0$, and therefore
\begin{align}
f^\beta(w) - f^\beta(w_*) \leq \frac{L}{2} \| w - w_* \|^2.
\end{align}
The result now follows from \cref{thm:strong-convex-thm}.

\subsection{Proof of \cref{thm:a-saga-converge-in-value}}

{\sloppy Based on \cref{lem:min_max_error_1}, it suffices to find a point at which
the value of $f^\beta$ is in the
\mbox{$\left(\epsilon - \frac{\log |\mathcal{Y}|}{2\beta}\right)$-neighbourhood} of
its optimal value.
Using \cref{lem:min_max_error_2}, we need
\begin{align}
\mathbb{E}\left[ f(w^t) - f(w_*) \right] \leq
\frac{L}{2}C_0\left(1-\frac{1}{\tau}\right)^t \leq \epsilon
- \frac{\log |\mathcal{Y}|}{2\beta} = \frac{\epsilon}{2}\,.
\end{align}
Following the same steps as in \cref{thm:asaga_complexity_1}, we conclude that
\begin{align}
t \geq \frac{\log \frac{2}{\epsilon} + \log \frac{C_0L}{2}}
{-\log\left( 1 - \frac{1}{\tau} \right)}\,.
\end{align}
Using the inequality $\log(1-x) \leq -x$, it suffices that
\begin{align}
\label{eq:qsaga_time_1}
t \geq \tau \left( \log \frac{2}{\epsilon} + \log \frac{C_0L}{2} \right).
\end{align}
From \eqref{eq:asaga_params}, we know that
\begin{align}
\tau =
\max\left\{ {2n}{},
\frac{2}{ \gamma\mu} \right\}
\leq \max\left\{ {2n}{},
\frac{68L}{ \mu} \right\},
\end{align}
where we have used the fact that $\theta \leq \frac{1}{\sqrt{D}}$.}

We recall that
${\max^\beta_{y \in \mathcal{Y}}} f_i(y, w)$
has Lipschitz continuous gradients with parameter $\beta D \Delta^2 + \ell$
(see \cite{beck2012smoothing}), so the function $f^\beta$ has
Lipschitz continuous gradients with parameter $L = \beta D \Delta^2 + \ell$.
Hence,
\begin{align}
\tau
\leq \max\left\{ {2n}{},
\frac{68(\beta D \Delta^2 + \ell)}{ \mu} \right\}.
\end{align}
Since $\beta = \frac{2\log |\mathcal{Y}|}{\epsilon}$, for sufficiently small
$\epsilon$, the second term dominates and we have
\begin{align}
\tau \leq \frac{68(\beta D \Delta^2 + \ell)}{ \mu}\,.
\end{align}
Replacing the values of $L$, $\mu$, and $\tau$ in the formulae, we get
\begin{align}
t \geq \frac{68(\frac{2\log |\mathcal{Y}|}{\epsilon} D \Delta^2 + \ell)}{ \mu}
\left(\log \frac{2}{\epsilon} + \log \frac{C_0 L}{2} \right).
\end{align}
Note that $C_0L= O(\frac{L}c)= O(n)$, so the time complexity
is $t= O\left((\frac{D \Delta^2\log |\mathcal{Y}|}{ \mu\epsilon}  + \frac{\ell}{ \mu})
\left(\log \frac{1}{\epsilon} + \log n \right)\right)$,
proving the claim.

\subsection{Proof of \cref{lem:qsaga_grad}}

Each iteration of SAGA requires finding all partial derivatives of $f_i$ for a
random choice of $i$ with precision $\theta$. Since $\epsilon$ is small, based
on \eqref{eq:asaga_params}, we have
$\theta = O\left(\frac{1}{\sqrt{D}}\frac{\mu \epsilon}
{\frac{1}{\beta D \Delta^2 + \ell}+\sqrt{\epsilon}}\right).$
We also note that $\Delta$,
which is a bound on the partial derivatives of ${\max_y} f_i(y,w)$,
is also a bound on the partial derivatives of
${\max_y}^\beta  f_i(y,w)$, because $\nabla_w {\max_y}^\beta f_i(y,w) =
\mathbb{E}(\nabla_w [ f_i(Y_i,w)])$. From \eqref{eq:smooth_ineq} we
know that $F \log |\mathcal{Y}|$ is a bound on $f^\beta_i$.
By replacing the  value of $\theta,F, \Delta$ in the cost of gradient calculation
the result follows.

\subsection{Proof of \cref{thm:saga-complexity-beta-and-original}}

To optimize $f^\beta$ using SAGA
with exact gradient evaluations, instead of the parameters from
\eqref{eq:asaga_params}, we set
\begin{align}
\begin{split}
\gamma = \frac{1}{2(\mu n + L)},
\quad
c = \frac{1}{2\gamma(1-\gamma \mu) n},
\quad
\alpha = \frac{2\mu n + L}{L},
\quad
\text{and}\quad \frac{1}{\tau} &= \gamma \mu
\end{split}
\end{align}
according to \cite{defazio2014saga}, with no assignment of $\theta$ (since there
are no additive errors after all). The rest of the proof follows the same steps
as in the proof of \cref{thm:asaga_complexity_1},
\cref{thm:a-saga-converge-in-value} and its corollary, we may optimize $f^\beta$
in order to estimate the optimal solution of $f$.

\subsection{Proof of \cref{thm:qsaga}}

Since $\epsilon$ is small, and $\beta$, $M$, and $\ell$ are fixed,
we can simplify the result of \cref{lem:qsaga_grad} and conclude that
each gradient could be estimated in
$O\left(\frac{D^{1.5} \beta F \sqrt{|\mathcal{Y}|} \log(|\mathcal{Y}|)\Delta}
{\mu \epsilon ({\beta D \Delta^2 + \ell})} \log \frac{D}{\zeta}\right)$
queries to the oracle for one of the $f_i$ and the same order of other quantum gates.
In $T$ iterations of Q-SAGA, if $\zeta = 1/(4T)$, the probability of
all gradient evaluations satisfying the additive
$\theta$ upper bound is larger than
$(1- \zeta)^{T} \geq 1- \left(\frac{1}{4T}\right){T} \geq \frac{3}4$.
The result follows from \cref{thm:asaga_complexity_1}.

\subsection{Proof of \cref{thm:qsaga-original}}

By replacing the value of $\beta$ from \cref{thm:a-saga-converge-in-value},
each gradient evaluation costs
\begin{align}
O\left(\left(\frac{1}{ D \Delta^2 \frac{\log|\mathcal{Y}|}{\epsilon} + \ell}+
\sqrt{\epsilon}\right)
\frac{D^{1.5} F \sqrt{|\mathcal{Y}|}
\log(|\mathcal{Y}|)\Delta}{ \mu \epsilon} \log \frac{D}{\zeta}\right)
\end{align}
queries to the oracle for one of the $f_i$ and the same order of other quantum gates
according to
\cref{lem:qsaga_grad}.
Using the fact that $\epsilon$ is small, this simplifies to
$O\left(
\frac{D^{1.5} F \sqrt{|\mathcal{Y}|} \log(|\mathcal{Y}|)\Delta}
{\mu \sqrt{\epsilon}} \log \frac{D}{\zeta}\right)$.
From \cref{thm:a-saga-converge-in-value}, we know that we need
$O\left((\frac{D \Delta^2\log |\mathcal{Y}|}{ \mu\epsilon}  + \frac{\ell}{ \mu})
\left(\log \frac{n}{\epsilon} \right)\right)$
gradient evaluations. Using the fact that $\epsilon$ is small, this
simplifies to
$O\left((\frac{D \Delta^2\log |\mathcal{Y}|}{ \mu\epsilon}  )
\left(\log \frac{n}{\epsilon} \right) \right)$.
By multiplying the number of gradient estimations with the
complexity of each, we get a total complexity of
\begin{align}
O\left(
\left(\frac{D^{1.5} F \sqrt{|\mathcal{Y}|}
\log(|\mathcal{Y}|)\Delta}{ \mu \sqrt{\epsilon}} \log \frac{D}{\zeta}\right)
\left(\frac{D\Delta^2\log |\mathcal{Y}|}{ \mu\epsilon} \right)
\left(\log \frac{n}{\epsilon} \right) \right).
\end{align}
As with the proof
of \cref{thm:qsaga} we should satisfy a failure probability of at most
$O(\frac{1}T)$ and get a total complexity of
\begin{align}
O\left(
\left(\frac{D^{1.5} F \sqrt{|\mathcal{Y}|}
\log(|\mathcal{Y}|)\Delta}{ \mu \sqrt{\epsilon}} \log D\right)
\left(\frac{D\Delta^2\log |\mathcal{Y}|}{ \mu\epsilon}\right)
\left(\log \frac{n}{\epsilon} \right)
\log \left(\frac{D\Delta^2\log |\mathcal{Y}|}{ \mu\epsilon}
\left(\log \frac{n}{\epsilon} \right)\right)
\right)
\end{align}
which, after simplification, completes the proof.

\end{document}